\pdfoutput=1
\documentclass[11pt]{article}

\usepackage{acl}
\usepackage{times}
\usepackage{latexsym}
\usepackage{amsmath} 
\usepackage{booktabs}
\usepackage{multirow}
\usepackage{graphics}
\usepackage{graphicx}
\usepackage{svg}
\usepackage{amsmath}

\usepackage[T1]{fontenc}
\usepackage[utf8]{inputenc}

\usepackage{microtype}

\usepackage{inconsolata}

\newcommand{\hide}[1]{}

\title{\textsc{Slang}: New Concept Comprehension of Large Language Models}

\author{Lingrui Mei$^{1,4}$ \quad Shenghua Liu$^{1,4}$\thanks{ Corresponding author.}\quad  Yiwei Wang$^{2,3}$ \quad Baolong Bi$^{1,4}$ \quad Xueqi Cheng$^{1,4}$ \\
	$^1$CAS Key Laboratory of AI Safety, Institute of Computing Technology, CAS \\
	$^2$University of California, Los Angeles \quad $^3$University of California, Merced \\
        $^4$University of Chinese Academy of Sciences \\
        \texttt{ \small{meilingrui22@mails.ucas.ac.cn}}
        \texttt{  \small{wangyw.evan@gmail.com}}
	\texttt{\small{\{liushenghua,bibaolong23z,cxq\}@ict.ac.cn}}
 }

\begin{document}
\maketitle
\begin{abstract}
The dynamic nature of language, particularly evident in the realm of slang and memes on the Internet, poses serious challenges to the adaptability of Large Language Models (LLMs). 
Traditionally anchored to static datasets, these models often struggle to keep up with the rapid linguistic evolution characteristic of online communities. 
This research aims to bridge this gap by enhancing LLMs' comprehension of the evolving new concepts on the Internet, without the high cost of continual retraining. 
In pursuit of this goal, we introduce \textbf{\textsc{Slang}}, a benchmark designed to autonomously integrate novel data and assess LLMs' ability to comprehend emerging concepts, alongside \textbf{\textsc{Focus}}, an approach uses causal inference to enhance LLMs to understand new phrases and their colloquial context. 
Our benchmark and approach involves understanding real-world instances of linguistic shifts, serving as contextual beacons, to form more precise and contextually relevant connections between newly emerging expressions and their meanings. 
The empirical analysis shows that our causal inference-based approach outperforms the baseline methods in terms of precision and relevance in the comprehension of Internet slang and memes. \footnote{Our code is available at \url{https://github.com/Meirtz/FocusOnSlang-Toolbox}.}
% \footnote{Our code is available at \url{https://anonymous.4open.science/r/FocusOnSlang-Toolbox-92C7}.}

\end{abstract}

\section{Introduction}
% The evolution of human language mirrors our capacity for innovation and adaptation, evolving from simple early forms to the complex structures observed in contemporary society.
% Recently, the language evolution has been accelerated by the developing communication technology, particularly the Internet, which has introduced new dimensions to linguistic evolution (\citealp{oxfordhb}; \citealp{firth2019online}; \citealp{hammarstrom2016linguistic}). 
Recently, language evolution has been accelerated by the online community, which has introduced new dimensions to linguistic shifts (\citealp{oxfordhb}; \citealp{firth2019online}; \citealp{hammarstrom2016linguistic}). 
These rapid changes in language pose serious challenges to the Large Language Models (LLMs) on understanding the newly emerging concepts (\citealp{yang2023out}; \citealp{sun-etal-2021-computational}).

Generally, LLMs are trained on static data (\citealp{brown2020language}), which limits their adaptivity to the dynamic and ever-evolving nature of human language. 
This limitation is particularly pronounced in the context of digital communication, where new forms of expression and concepts emerge at an unprecedented pace (\citealp{sun-etal-2021-computational}). 
% Hence, it is critical to evaluate the capability of LLMs to adapt to and grasp these linguistic changes and novel concepts without the necessity for constant retraining, fine-tuning with newer datasets, or accessing external information.
Hence, it is essential for LLMs to understand linguistic shifts and new concepts without constant updates or external data.

\begin{figure}
    \centering
    \includegraphics[width=1\linewidth]{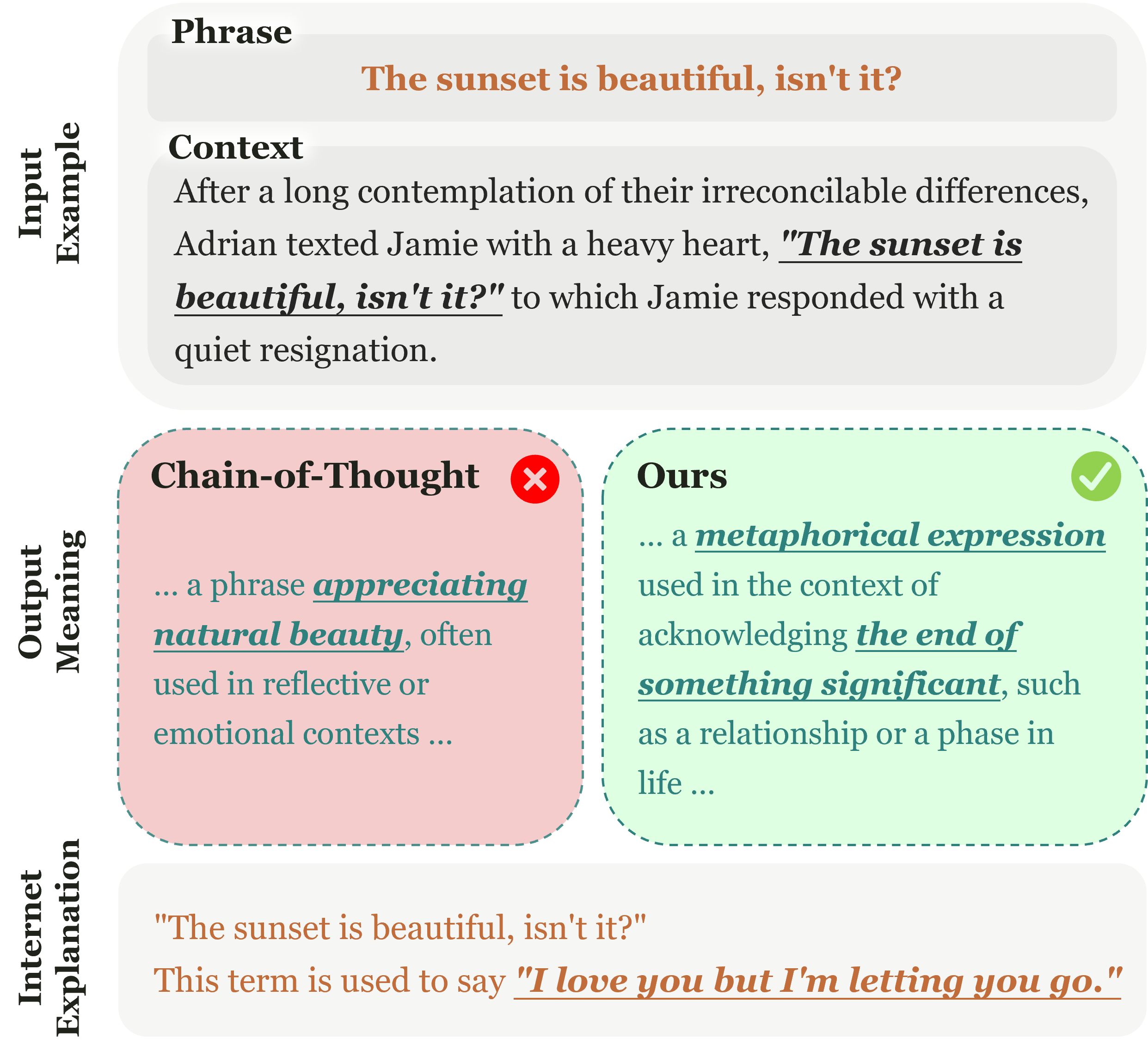}
    \caption{Comparative analysis of LLMs' understanding of new phrases using CoT \cite{wei2022chain} and \textsc{Focus} methods. The left side demonstrates the limited understanding through the CoT approach, focusing on the literal interpretation. In contrast, the right side using the \textsc{Focus} method shows the model's enhanced capability to grasp metaphors and deeper meanings.}
    \label{fig:example}
    \vspace{-10pt}
\end{figure}

Moreover, LLMs often make decisions based on superficial patterns rather than justified reasons. This can hinder their ability to accurately interpret and follow human instructions, as highlighted in several studies (\citealp{tang-etal-2023-large}; \citealp{wang-etal-2022-rely}, \citealp{zhou-etal-2023-context}, \citealp{wang-etal-2023-causal}). For example, as depicted in Figure \ref{fig:example}, the Chain-of-Thought (CoT) prompting \cite{wei2022chain} simply interprets the phrase \textit{The sunset is beautiful, isn't it?} and misses the deeper, metaphorical meaning, which could represent an acknowledgment of an ending, like concluding a life phase or a relationship, in a complex conversation. This situation underscores the importance of enhancing and evaluating LLMs in a way that goes beyond their performance metrics. It's essential to consider the fundamental principles that guide their decision-making processes.

Therefore, we propose \textbf{\textsc{Slang}} (\textbf{S}imilarity of \textbf{L}exical \textbf{A}nalysis a\textbf{N}d \textbf{G}rasp), a benchmark to assess language models' adaptability to linguistic shifts, and \textbf{\textsc{Focus}} (\textbf{F}actual c\textbf{O}ntext \textbf{C}a\textbf{U}sal analysi\textbf{S}), an approach based on causal inference for enhancing comprehension of new concepts. 

The \textsc{Slang} benchmark, developed from UrbanDictionary \cite{urbandictionary}, focuses on assessing the capability of language models to maintain coherence and accuracy in the face of dynamic and unconventional language use, such as slang and idiomatic expressions, thereby evaluating the capability of LLMs in grasping new concepts. 
We select recent entries after a specific cutoff date and filter out phrases already likely in LLM training data. We utilize user-generated ratings (upvotes and downvotes) to refine the dataset, ensuring its quality and comprehensiveness. The dataset is then standardized into a formal dictionary format, simplifying explanations and examples for universal understanding while retaining original meanings. This preprocessing approach ensures \textsc{Slang} effectively evaluates LLMs' adaptability to linguistic shifts.

% \textsc{Slang} benchmark focuses on assessing the capability of language models to maintain coherence and accuracy in the face of dynamic and unconventional language use, such as slang and idiomatic expressions, thereby evaluating the capability of LLMs in grasping new concepts. We constructs \textsc{Slang} from UrbanDictionary, featuring a spectrum of factual and counterfactual scenarios. 

\textsc{Focus} employs causal inference to enhance models' comprehension of new concepts within evolving linguistic contexts. By analyzing causal relationships in language, \textsc{Focus} advances models' predictive capabilities beyond traditional correlation-based learning. This method allows for a nuanced grasp of language dynamics, improving models' adaptability and effectiveness in applications requiring deep understanding of language use.
\textsc{Focus} significantly enhanced performance in language model comprehension, demonstrating superior precision and adaptability. With Claude 3, \textsc{Focus} achieved an $\text{F}_1$ score of 0.4596, precision of 0.4452, and recall of 0.4827, alongside an accuracy of 89.7\%, outperforming previous methods in comprehension and adaptability.
%This underscores \textsc{Focus}'s impact in enhancing language model comprehension and adaptability to linguistic shifts, showcasing its applicability in new concepts understanding.

%\textsc{Focus} delves into the causal relationships in language, particularly in evolving linguistic contexts, aiming to enhance the precision and adaptability of language models, and enrich models' grasp of diverse new concepts. 

The codes for the \textsc{Slang} and \textsc{Focus} toolboxes are open-sourced, contributing to the community's resources for advancing language model development.

% \reminder{Results and open source code }

\section{\textsc{Slang}}

We introduce \textsc{Slang} in response to rapidly evolving language. \textsc{Slang} benchmark evaluates LLMs' capability to interpret the dynamic landscape of user-generated new concepts. It uniquely features factual and counterfactual datasets, each crucial for gauging LLM adaptability. We detail \textsc{Slang}'s dataset construction, and evaluation metrics in the following subsections.
%This structure ensures a comprehensive assessment of LLMs' capabilities in navigating the complexities of modern linguistic trends.

%\subsection{Problem}

%In the dynamic field of digital communication, slang and idiomatic expressions evolve rapidly. This evolution poses significant challenges for large language models (LLMs). Our study focuses on the probability $P(Y|W,X;\mathcal{M})$. This probability represents the likelihood that a given explanation $Y$ is derived from a phrase $W$ within its context $X$, as interpreted by the model $\mathcal{M}$.

%We extend this inquiry to encompass conditional dependencies, considering $P(Y|W,X;\mathcal{M}) = \sum_{i} P(Y|W,X,E_i;\mathcal{M})P(E_i|X)$, where $E_i$ represents various contextual entities influencing $W$ within $X$. This comprehensive approach underscores the multifaceted nature of language interpretation, casting our model as both a decipherer and a storyteller in this ever-evolving linguistic landscape.

\subsection{Preprocessing}\label{subsec:dataset}
\paragraph{Extraction}
% In this research, our primary focus revolves around the development of a comprehensive dataset derived from UrbanDictionary, a platform renowned for its user-contributed definitions, which encompass slang, phrases, and cultural idioms. This resource proved invaluable in extracting critical elements, including the phrases, their definitions, usage examples, and user feedback in the form of upvotes and downvotes.

Our dataset construction involved extracting numerous concepts from UrbanDictionary, a platform known for user-generated content that reflects current language trends and the evolving internet lexicon, making it a unique, constantly updated repository and dynamic forum for new Internet language concepts.
%In the construction of the dataset, our approach involved the systematic extraction of a substantial number of concepts from UrbanDictionary \cite{urbandictionary}, a widely recognized and reliable online repository of Internet slang and colloquial terms. UrbanDictionary stands out for its user-driven content, reflecting current language trends and the evolving nature of the internet lexicon. This platform serves as both a repository and a dynamic community, continuously enriching the evolving landscape of internet language with new concepts. 
Specifically, our approach involved selecting concepts added after a predetermined date to ensure content novelty. We meticulously extract relevant data, including the phrase, its user-provided definitions, usage examples, and user-generated ratings (\emph{upvotes} and \emph{downvotes}). 
%This approach ensured that our dataset was grounded in authentic user-driven language trends. 
Additionally, the data construction pipeline is set up to automatically include fresh, non-member data for upcoming cutoff dates, ensuring the dataset stays up-to-date and comprehensive.

\paragraph{Filtering}

\begin{figure}
    \centering
    \includegraphics[width=1\linewidth]{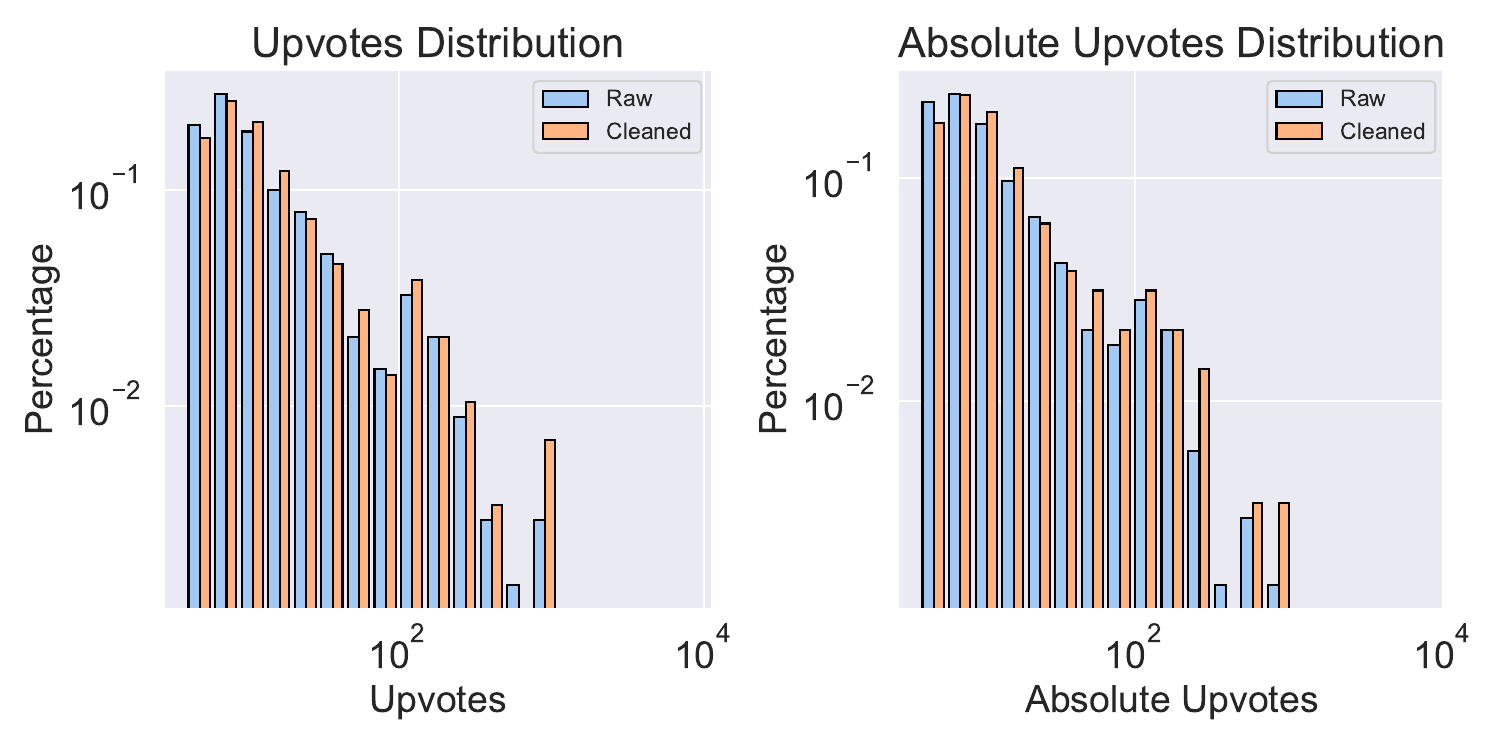}
    \caption{Comparative histograms illustrating the distribution of \emph{upvotes} (left) and \emph{absolute upvotes} (right) across dataset entries. Both histograms are plotted on a logarithmic scale with the vertical axis representing the log percentage of the total dataset and the horizontal axis indicating the log count of \emph{upvotes} or \emph{absolute upvotes}. The blue bars represent the raw data, while the orange bars depict the cleaned data, facilitating a direct comparison of the distributions before and after data cleaning.}
    \label{fig:vote_dist}
\end{figure}

The content filtering of our dataset consisted of several steps to ensure the quality and novelty of the concepts:

\begin{itemize}
    \item \textbf{Temporal Filtering:} 
    Many UrbanDictionary phrases are now common and potentially included in LLMs' training data. To ensure the novelty of our dataset, we leveraged the knowledge cut-off dates of LLMs, strategically selecting phrases that emerged after these dates. For instance, \texttt{gpt-4-0613} has a knowledge cut-off date of April 2021, as detailed in the OpenAI documentation\footnote{\url{https://platform.openai.com/docs/models}}, and we selected concepts that were added to UrbanDictionary after January 2022. This temporal gap was strategically chosen to include recent phrases that may have gained popularity online after the cut-off date but were not yet recorded in UrbanDictionary. 
    
    \item \textbf{User Rate Filtering:}
    We analyzed user-generated ratings to refine the dataset. Entries with overwhelmingly negative receptions (more than 80\% downvotes) were excluded. Comparative histograms in Figure \ref{fig:vote_dist} illustrate the distribution of \emph{upvotes} (left) and \emph{absolute upvotes} (right) across dataset entries before and after our cleaning process. This stringent data cleaning was crucial in ensuring the quality and reliability of the entries selected for our research. For details on the validation of user-generated votes, see Appendix \ref{appendix:validation_user_votes}. 
    
    \item \textbf{Removal of Inappropriate Content:}
    Inappropriate content, including NSFW material and hate speech, was removed to preserve academic integrity and ensure quality. 
    
    \item \textbf{Novelty Check:}
    To ensure that the concepts in our dataset were unknown to the LLMs, we employed \texttt{gpt-4-0613} for thorough filtration based on the method described by \citealp{yin-etal-2023-large}. Additionally, due to the existence of models with different cut-off dates, we applied the "\texttt{needle in a haystack}" test (\citealp{Kamradt2023}, see Appendix \ref{appendix:needle_in_a_haysatck}) to confirm the novelty of the knowledge. This meticulous validation process involved strategically embedding the selected phrases into extensive corpora and evaluating the models' capability to extract them.
\end{itemize}
After the temporal filtering step, we started with 7220 concepts. The subsequent steps filtered out 5463, 1328, and 21 samples respectively, resulting in a final dataset of 408 usable new concepts.

\paragraph{Factual dataset}
%Acknowledging the informal nature of the original dataset, we transformed the data into a uniform, formal dictionary format. This transformation involves simplifying the explanations and examples to make them universally understandable while preserving their original online meaning. The transformation process adhered to a custom-designed template, ensuring consistency and clarity across the concepts.
%We also enriched each explanation with four synonym-based variants, ensuring the capture of diverse potential responses.
%When evaluating predictions, we compare the model's output with all five versions of the transformed explanations (the original plus four synonym-augmented variants) and select the result that most closely matches. This method ensures a more comprehensive and accurate assessment of the model's capability to interpret and reproduce the factual content.

Following the above filtering steps, we obtained our factual dataset. Acknowledging the informal nature of the original data, we transformed it into a uniform, formal dictionary format (see Appendix \ref{sec:standardization}). This process involved simplifying explanations and examples for universal understanding while preserving their original meanings. We adhered to a custom-designed template for consistency and clarity across all concepts. Each explanation was enriched with four synonym-based variants to capture diverse potential responses. 

\paragraph{Counterfactual Dataset}

To further evaluate the ability of LLMs in understanding new concepts, we created the counterfactual dataset. This dataset is derived from the factual dataset by preserving the original phrases while modifying the contexts and explanations. Given the factual dataset $ \mathcal{D}_{\text{fact}} $, the generation process is structured as follows:

\begin{itemize}
    \item \textbf{Entity Extraction:} We extract the entities $ \mathbf{e} $ from each explanation $ \mathbf{y} $ in $ \mathcal{D}_{\text{fact}} $.
    \item \textbf{Counterfactual Replacement:} For each entity $ e_i \in \mathbf{e} $, we generate a counterfactual entity $ e_i' $ that is conceptually divergent from the original entity $ e_i $ while retaining the original phrase structure.
    \item \textbf{Context Construction:} Based on the counterfactual explanations $ y_i' $, we use GPT-4 to generate new contexts $ x_i' $. Consequently, the counterfactual dataset $ \mathcal{D}_{\text{cf}} $ is composed of pairs: $ \mathcal{D}_{\text{cf}} = \{(p_1, x_1', y_1'), (p_2, x_2', y_2'), \ldots, (p_n, x_n', y_n')\} $.
\end{itemize}

This approach ensures that while the original phrases are preserved, the contexts and explanations are transformed to convey entirely different meanings. Consequently, each entry in the counterfactual dataset introduces novel concepts that is distinct from the original dataset, providing unique challenges for the LLMs to interpret and understand.

\subsection{Metrics}
%Given the nature of this task as a question-answering (QA) problem, traditional QA metrics like F1 score, Recall, and Precision were employed \cite{yang-etal-2018-hotpotqa}. Moreover, we introduced BLEU (3-gram) \cite{papineni-etal-2002-bleu} and ROUGE \cite{lin-2004-rouge} as metrics to explore the capacity of the listed methods to generate higher quality interpretations with stricter similarity requirements. 
For this task, we employed traditional metrics like $\text{F}_1$ score, recall, and precision \cite{yang-etal-2018-hotpotqa}, and added BLEU (3-gram) \cite{papineni-etal-2002-bleu} and ROUGE \cite{lin-2004-rouge} for stricter quality checks.
Considering that language models might output synonymous interpretations with varied wording, we also incorporated sentence similarity measures such as sentence-level similarity and SimCSE \cite{gao-etal-2021-simcse}, and sentence similarity \cite{sensim-v2} as \emph{Similarity} by calculating the cosine similarity of the embeddings of  \texttt{all-mpnet-base-v2} (\citealp{sensim-v2}; \citealp{song2020mpnet}). A positive sample was considered for accuracy calculation if its SimCSE score exceeded 0.7. 
We also generated five lexically varied yet syntactically and semantically identical interpretations for each dataset entry. To determine the final metric, we selected the interpretation with the highest BLEU score.

\section{\textsc{Focus}}

This section outlines the novel approach employed in our research to enhance the adaptability of LLMs in understanding the evolving human language. 
Our method analogizes the dynamic nature of language to a continuously evolving entity that requires adaptive comprehension strategies.

\begin{figure}
    \centering
    \includegraphics[width=1\linewidth]{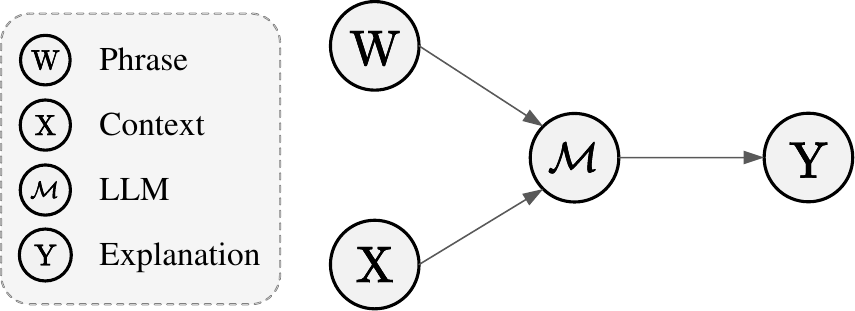}
    \caption{Structural Causal Model (SCM) of LLMs for interpreting new phrases, excluding confounders. The variables $X$ and $W$ encapsulate users' complex intentions and thoughts, which span intricate emotional expressions, cultural insights, and extensive internet-specific knowledge. Grasping these nuanced aspects directly through an LLM is a challenging endeavor.}
    \label{fig:scm1}
    \vspace{0pt}
\end{figure}

\subsection{Causal Analysis}

\begin{figure*}
    \centering
    \includegraphics[width=0.9\linewidth]{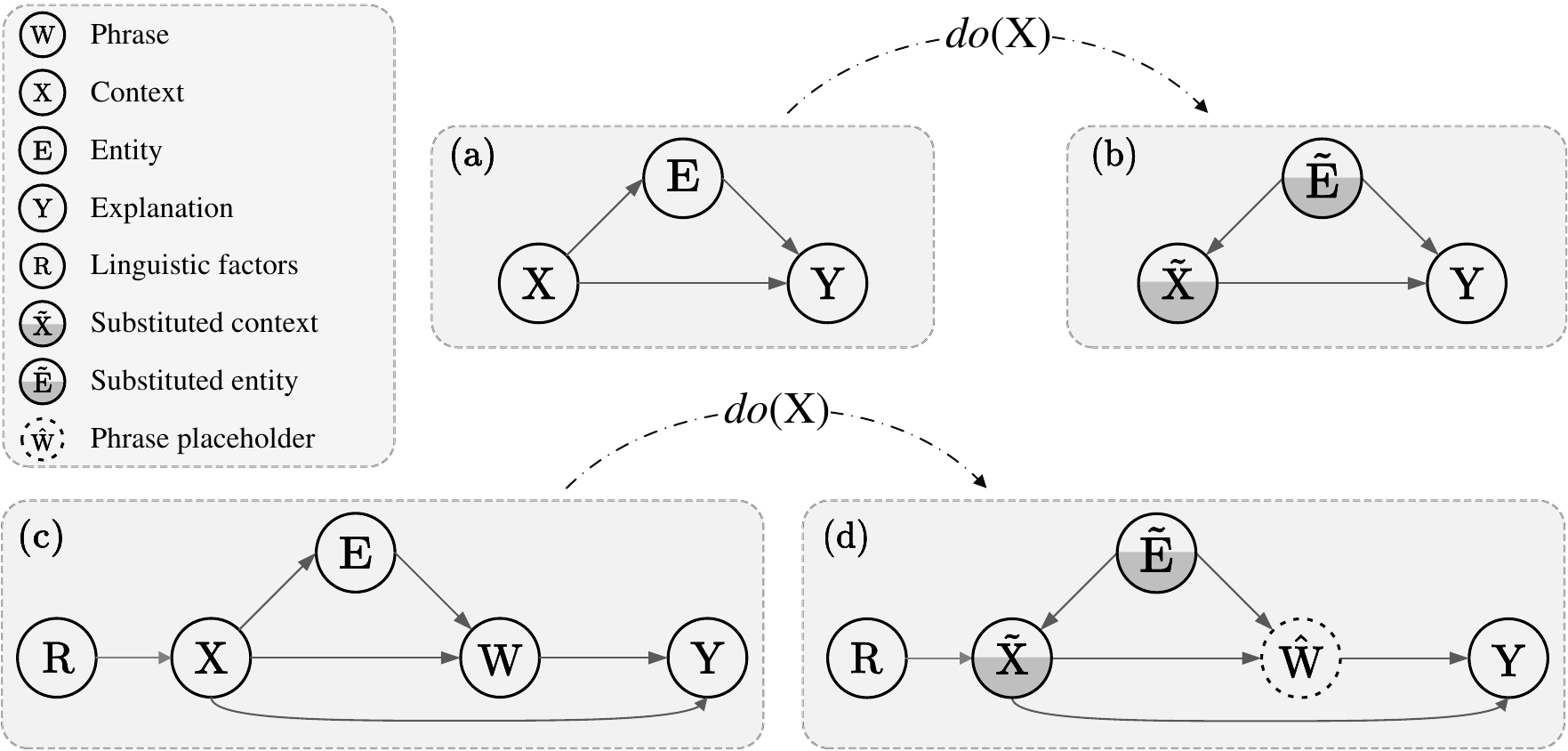}
    \caption{SCM analysis in \textsc{Focus} methodology. This figure presents the complex process of causal inference in the understanding of new phrases, highlighting how the \textsc{Focus} approach systematically analyzes and interprets intricate language patterns, emphasizing the causal links between linguistic elements and their interpretive outcomes.}
\label{fig:scm2}
\vspace{-5pt}
\end{figure*}

Structural Causal Models (SCMs) serve as vital tools for elucidating the relationships and influence pathways among variables. 
We propose a simplified SCM (see Figure \ref{fig:scm1}) to delineate the interpretative processes of an LLM when confronted with novel phrases within their context. 
In this model, users supply both the phrase $W$ and its context $X$, which are then inputted into the LLM, represented by the direct links $X \rightarrow \mathcal{M} \rightarrow Y $ and $W \rightarrow \mathcal{M} \rightarrow Y$, where $\mathcal{M}$ denotes the LLM and $Y$ is the output explanation. 
These causal links, free from confounders, capture the logical chain from input to interpretation. 
This SCM sets the stage for our forthcoming discussion where we reintroduce and scrutinize these confounders, thus laying the groundwork for a comprehensive causal analysis.

\paragraph{Analysis of entity}
Initially, referencing a typical SCM framework (\citealp{wang-etal-2023-causal}; \citealp{wang-etal-2022-rely}), we assume SCM $S=\{X,E,Y\}$, where $X$ denotes context/input, and $E$ represents confounder, including phrases and other entities within the context. From the perspective of human understanding of new phrases, the interpretation $Y$ is derived from the context $X$ and its entities $E$. The confounder $E$ can be extracted from $X$, leading to the relationships $X \rightarrow Y \leftarrow E $ and $X \rightarrow E $ in our model.  By applying the $do$-operation \cite{judea1990equivalence}, which denoted as $do(X)$, we follow the guidelines (\citealp{wang-etal-2023-causal}; \citealp{wang-etal-2022-rely}) to conduct a rigorous causal inference, ensuring that the main effects of the textual context are captured without losing entity information. This operation aims to isolate the effect of the context $X$ on the confounder $E$. Consequently, the relationship between $X$ and $E$ undergoes a change, becoming $\tilde X \leftarrow \tilde E$, where $\tilde{E}$ represents the modified entity, and $\tilde{X}$ denotes the context obtained after the $do$ operation which effectively substitutes the actual entity.

\paragraph{Analysis of linguistic factors}
In our SCM, denoted as $S=\{X,W,E,Y,R\}$, the output variable $Y$, as an endogenous variable, is influenced by the input context $X$, the input phrase $W$, other entities $E$, and finally, linguistic factors $R$, which include the linguistic structure, style, theme, and cultural background, crucially shape $X$ and constitute the exogenous variables. The model includes direct paths $X \rightarrow Y$ and $W \rightarrow Y$, indicating the immediate influence of context and phrase on the interpretation. The backdoor path $E \leftarrow X \rightarrow W$ and the indirect path $X \rightarrow E \rightarrow W$ demonstrate the mediated effects. The path $R \rightarrow X$ highlights the exogenous influence of linguistic factors on context. 
% This configuration captures the causal mechanisms underlying natural language understanding, emphasizing the direct, indirect, and mediated relationships among the variables. 
% This configuration emphasizes the direct, indirect, and mediated relationships among the variables. 
The equation for the causal effect in this context is as follows:
\begin{align*}
& P(Y=y |X=x) \\ =  & \sum_{e,r} P(Y=y | X=x, W=w, E=e, R=r) \\ & P(W=w|X=x, E=e, R=r)\\ & P(E=e) P(R=r)
\end{align*}
In this formula, $P(Y=y | X=x)$ represents the probability of the outcome variable $Y$ being a particular value $y$, given the context $X$ is set to $x$. The summation over $e$ and $r$ encompasses all possible combinations of the values of the entities $E$ and linguistic factors $R$. For each combination, the formula calculates the conditional probability of $Y=y$ given $X=x$, the specific phrase $W=w$, the entity $E=e$, and the linguistic factor $R=r$. This conditional probability is further modulated by the probabilities $P(W=w|X=x, E=e, R=r)$, $P(E=e)$, and $P(R=r)$. These terms represent the likelihood of observing the phrase $W=w$ conditioned on the context, entities, and linguistic factors, as well as the inherent probabilities of the entities $E=e$ and the linguistic factors $R=r$. This comprehensive approach allows for a nuanced understanding of how context, entities, and linguistic factors collectively influence the interpretation $Y$.

\subsection{Context-based Causal Intervention}

Subsequently, we refine our causal intervention approach in the SCM, focusing on the role of context and linguistic factors. To concentrate on the contextual content and reduce reliance on shortcuts, $E$ is replaced with $\tilde{E}$, consequently transforming $X$ into $\tilde{X}$, (as illustrated in Figure \ref{fig:scm2} (a) and (b)). Further, to eliminate the entity bias in $W$, it is replaced with a specific placeholder, denoted as $\hat{W}$ (as illustrated in Figure \ref{fig:scm2} (c) and (d)). This alteration aids in isolating the effect of $W$ while excluding entity-specific biases. Overall, these adjustments accentuate the role of $R$ in shaping the context, enhancing the model's capacity to highlight the influence of linguistic factors in a bias-free manner. The updated causal effect formula in the SCM is thus:
\begin{align*}
& P(Y=y | do(X=x)) \\ 
=  & \sum_{\tilde{e},r} P(Y=y | X=x, \tilde{W}=\tilde{w}, \tilde{E}=\tilde{e}, R=r) \\ 
& P(\tilde{W}=\tilde{w}| do(X = x), \tilde{E}=\tilde{e}, R=r) \\
& P(\tilde{E}=\tilde{e}) P(R=r) 
\\ =  & \sum_{\tilde{e},r} P(Y=y | X=x, \tilde{W}=\tilde{w}, \tilde{E}=\tilde{e}, R=r) \\ 
& P(\tilde{W}=\tilde{w}|\tilde{E}=\tilde{e}, R=r) P(\tilde{E}=\tilde{e}) P(R=r) 
\end{align*}
This revised formula ensure a more comprehensive and nuanced analysis of the causal dynamics within the SCM, post-intervention. This approach ensures a more robust and bias-free interpretation within the SCM framework. This equation accounts for the altered relationships in the SCM after $do$-operation. The conditional probabilities and summations are now over the new variables $\tilde{E}$ and $\tilde{W}$, while maintaining the original structure's intent to adjust for confounding effects and capture the influence of modified entities in the context $X$. 

\begin{figure*}
    \centering
    \includegraphics[width=1\linewidth]{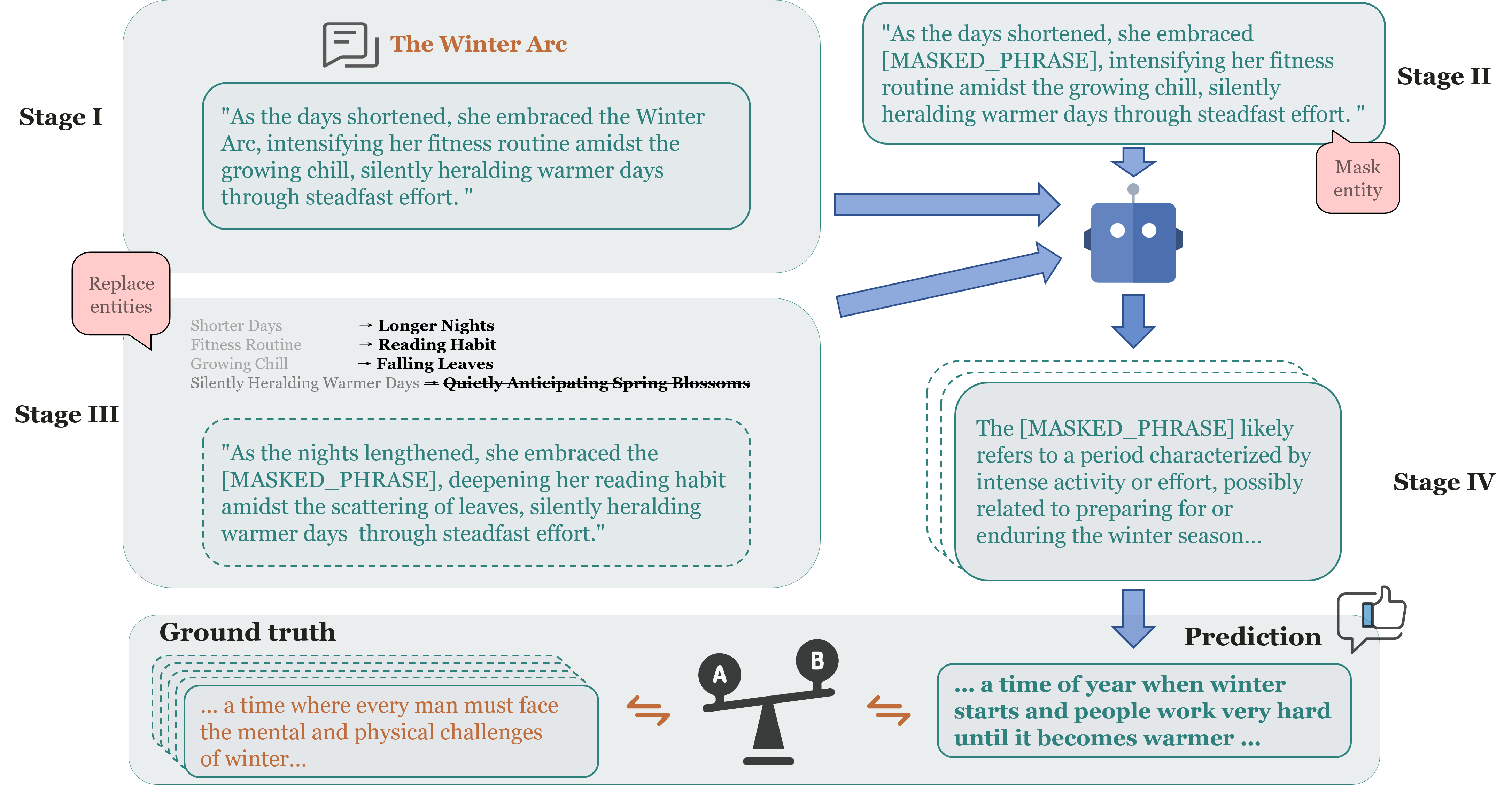}
    \caption{The four-stage pipeline of \textsc{Focus}.}
    \label{fig:pipeline}
    \vspace{-10pt}
\end{figure*}

The core idea of \textsc{Focus} is to explore how language models can adhere to guidelines to better understand the content of the context. The goal is to enable LLMs to analyze phrases according to usage examples and provide counterfactual interpretations, thereby understanding the evolving semantics of language. To achieve this objective, we propose a four-stage method (as shown in  Figure \ref{fig:pipeline}), each with its principles:

%\paragraph{Direct Inquiry (DI)} The DI stage is the foundational step in our \textsc{Focus} approach. It primarily involves inputting the usage example (context) $X$ and the phrase $W$ directly into the language model. This stage is represented as $Y_{DI} = \mathcal M (X, W, E)$, where $\mathcal M$ is the function that defines the processing of the language model. The aim here is to assess the direct effect of the context $X$ on the inferred meaning $Y$ in a straightforward and intervention-free manner. This allows the model to naturally derive the meaning within the given context. Special emphasis is placed on maintaining the integrity of all confounders, ensuring that the phrase's morphology, literal meaning, and context-based interpretation are coherently analyzed. The DI stage sets the baseline for understanding the meaning of the phrase before any further analytical layers are added in the subsequent stages.

\paragraph{Direct Inquiry (DI)} In the Direct Inquiry (DI) stage, we input the usage example (context) $X$ and phrase $W$ into the language model. Represented as $Y_{DI} = \mathcal M (X, W, E)$, this stage aims to evaluate the direct effect of $X$ on the inferred meaning $Y$. It allows the model to derive meaning naturally within the given context, focusing on the coherence of the phrase's morphology, literal meaning, and context-based interpretation. DI sets a baseline for understanding the phrase's meaning before adding more analytical layers.

%\paragraph{Masked Entity Inquiry (MEI)} The MEI stage is a crucial component of our \textsc{Focus} methodology, designed to mitigate the impact of the phrase's literal meaning on its interpretation. In this stage, the phrase $W$ is masked within the context $X$, leading to an evaluation of the meaning $Y$ in the absence of direct influence from $W$. This process is denoted as $Y_{MEA}, \hat{W}_{MEA} = \mathcal{M}(X_{\text{masked}}, E)$, where $\mathcal{M}$ represents the language model. The primary objective is to understand the meaning based on context, prompting the model to propose synonyms or near-synonyms that could feasibly replace the masked phrase. The MEI stage allows for a nuanced analysis of how the interpretation $Y$ is influenced when $W$ is obscured, yet still impacted by the entities $E$. This stage is instrumental in mitigating the literal interpretation bias of $W$, focusing instead on deriving meaning from the broader linguistic factors present in $X$ and $E$.

\paragraph{Masked Entity Inquiry (MEI)} In the Masked Entity Inquiry (MEI) stage, we mask the phrase $W$ within context $X$ to analyze the meaning $Y$ without $W$'s direct influence. This process, $Y_{MEA}, \hat{W}_{MEA} = \mathcal{M}(X_{\text{masked}}, E)$, helps the model suggest synonyms or near-synonyms for the masked phrase. MEI focuses on extracting meaning from broader linguistic factors in $X$ and $E$, reducing bias towards $W$'s literal interpretation and enhancing context-based understanding.

\begin{table*}[ht]
    \centering
    \resizebox{2\columnwidth}{!}{
    \begin{tabular}{llcccccccccc}
    \toprule
    \multirow{2}{*}{\textbf{Model}} & \multirow{2}{*}{\textbf{Prompting Method}} & \multirow{2}{*}{$\textbf{F}_\textbf{1}$} & \multirow{2}{*}{\textbf{Precision}} & \multirow{2}{*}{\textbf{Recall}} & \multirow{2}{*}{\textbf{BLUE}} & \multicolumn{3}{c}{\textbf{ROUGE}} & \multirow{2}{*} {\textbf{Similarity}}  & \multirow{2}{*}{\textbf{SimCSE}} &\multirow{2}{*}{\textbf{ACC (\%)}} \\ 
    \cline{7-9} 
    \multicolumn{6}{c}{} &  1 & 2 & \textit{L} & & \\
    % \cline{6-8} \\
    \midrule
    \multirow{4}{*}{\textbf{Mistral-7B}}
    & Direct \cite{NEURIPS2022_b1efde53} & 0.1869 & 0.2655 & 0.1684 & 0.0372 & 0.2078 & 0.0481 & 0.1452 & 0.1507 & 0.6084 & 43.1 \\
    & CoT \cite{wei2022chain} & 0.3453 & 0.3227 & 0.3706 & 0.1978 & 0.3657 & 0.1602 & 0.2744 & 0.4713 & 0.7132 & 68.5 \\
    & CauView \cite{wang-etal-2023-causal} & 0.3012 & 0.2787 & 0.3384 & 0.1561 & 0.3272 & 0.1336 & 0.2449 & 0.4862 & 0.7127 & 64.3 \\
    & \textbf{\textsc{Focus} (Ours)} & \textbf{0.3703} & \textbf{0.3555} & \textbf{0.3878} & \textbf{0.2493} & \textbf{0.3942} & \textbf{0.1866} & \textbf{0.3011} & \textbf{0.5121} & \textbf{0.7469} & \textbf{76.0} \\
    
    \midrule
    \multirow{4}{*}{\textbf{GPT-4}} %& Base & -- & {\phz}61.0 & 30.0 & 36.6\\
    % \cline{2-6}

    & Direct \cite{NEURIPS2022_b1efde53} & 0.2308 & 0.3474 & 0.1917 & 0.0483 & 0.2597 & 0.0606 & 0.1859 & 0.1616 & 0.6476 & 47.2 \\
    & CoT \cite{wei2022chain} & 0.4123 & 0.3947 & 0.4244 & 0.2384 & 0.4370 & 0.1927 & 0.3299 & 0.5521 & 0.7883 & 79.3 \\
    & CauView \cite{wang-etal-2023-causal} & 0.3602 & 0.3444 & 0.3948 & 0.1987 & 0.3917 & 0.1643 & 0.3032 & 0.5515 & 0.7636 & 74.7 \\

    & \textbf{\textsc{Focus} (Ours)} & \textbf{0.4446} & \textbf{0.4280} & \textbf{0.4714} & \textbf{0.3177} & \textbf{0.4721} & \textbf{0.2332} & \textbf{0.3652} & \textbf{0.6032} & \textbf{0.8216} & \textbf{88.2} \\

    \midrule
    \multirow{4}{*}{\textbf{Claude 3}}
    & Direct \cite{NEURIPS2022_b1efde53} & 0.2395 & 0.3538 & 0.2171 & 0.0552 & 0.2645 & 0.0673 & 0.1904 & 0.1947 & 0.6714 & 51.4 \\
    & CoT \cite{wei2022chain} & 0.4276 & 0.4082 & 0.4606 & 0.2471 & 0.4492 & 0.2014 & 0.3371 & 0.5628 & 0.7948 & 81.2 \\
    & CauView \cite{wang-etal-2023-causal} & 0.3752 & 0.3550 & 0.4229 & 0.2015 & 0.4035 & 0.1707 & 0.3041 & 0.5778 & 0.7932 & 76.6 \\
    & \textbf{\textsc{Focus} (Ours)} & \textbf{0.4596} & \textbf{0.4452} & \textbf{0.4827} & \textbf{0.3264} & \textbf{0.4835} & \textbf{0.2373} & \textbf{0.3729} & \textbf{0.6109} & \textbf{0.8354} & \textbf{89.7} \\
    
    \bottomrule
    \end{tabular}
    }
    \caption{Performance results on the factual dataset. For results on more models, see Appendix \ref{appendix:new_factual_results}.}
    \label{table:factual_results}

\end{table*}

%\paragraph{Entity Replacement Inquiry (ERI)} The ERI stage plays a pivotal role in our \textsc{Focus} methodology, where we explore the impact of varying entity relationships on the interpretation of phrases to delve deeper into context-based interpretation. In this stage, we implement an entity replacement strategy combined with a dropout rate, selectively altering entities within the context $X$. This process, denoted as $Y_{ERI}, \hat{W}_{ERI} = \mathcal{M}(\tilde X_{\text{masked}}, \tilde E_{\text{replaced}})$, involves generating multiple context sentences with replaced entities, while a dropout rate determines the extent of entity replacement. The addition of this dropout strategy is not only to introduce randomness but also to prevent the emergence of new entity biases that could arise from completely replacing all entities. By maintaining a mix of original and altered entities, we ensure the robustness of understanding while mitigating the risk of introducing new biases. The language model $\mathcal{M}$ then conducts multiple inferences on these varied contexts to synthesize a cohesive interpretation. This approach enhances the robustness of the model by considering a diverse range of entity interactions. The ERI stage is fundamental in understanding the nuanced effects of entity dynamics within different contexts, providing a more comprehensive causal analysis of phrase interpretation.

\paragraph{Entity Replacement Inquiry (ERI)} In the Entity Replacement Inquiry (ERI) stage, we alter entities in context $X$ to assess phrase interpretation variability. We use a dropout rate for entity alteration in $Y_{ERI}, \hat{W}_{ERI} = \mathcal{M}(\tilde X_{\text{masked}}, \tilde E_{\text{replaced}})$. This introduces a balance of original and new entities, enhancing model robustness without bias. ERI helps understand entity dynamics' effects on interpretation, providing deeper causal analysis.

%\paragraph{Synthesis (SY)} The SY stage is the integrative phase of the \textsc{Focus} methodology, symbolized as $Y_{FS} = \mathcal{M}(Y_{DI}, Y_{MEA}, Y_{ERI})$. This stage synthesizes insights from Direct Inquiry (DI), Masked Entity Analysis (MEA), and Entity Replacement Inquiry (ERI). It evaluates the interplay between direct, contextually-driven, and variable entity interpretations to forge a comprehensive understanding of the phrase $W$ within context $X$ and its relation to entities $E$. In FS, we assess causal relationships, reconciling the varied interpretations and confounding factors identified in previous stages. This synthesis refines our understanding of language model interpretations, emphasizing contextual richness and linguistic nuance. The SY stage thus concludes the \textsc{Focus} methodology, offering a multi-dimensional perspective on language model analytics.

\paragraph{Synthesis (SY)} In the Synthesis (SY) stage, we integrate insights from Direct Inquiry, Masked Entity Inquiry, and Entity Replacement Inquiry. Represented as $Y_{FS} = \mathcal{M}(Y_{DI}, Y_{MEA}, Y_{ERI})$, this phase evaluates the interplay between direct, contextual, and entity-variable interpretations. SY reconciles varied interpretations and confounders, refining the model's understanding of language nuances. This final stage offers a multi-dimensional perspective on language model analytics, emphasizing contextual richness.

\section{Experiments}

\subsection{Setup}
\paragraph{Large Language Models} Following the preprocessing method described in Section \ref{subsec:dataset}, we filtered our initial dataset of 7220 concepts, resulting in 408 new concepts. These evaluations were conducted using Claude 3, GPT-4, Mistral-7B, and other popular models. For detailed experimental setup and additional results for other models, please refer to Appendix \ref{appendix:setup} and Appendix \ref{appendix:additional_results}.

 %These models were selected for their advanced capabilities and the relevance of their knowledge bases, which have a cut-off date prior to April 2021. As we discussed in \ref{subsec:dataset}, this cut-off is crucial to ensure the novelty and relevancy of our dataset, as it aligns with the temporal frame of our selected UrbanDictionary concepts. 
%Moreover, the choice of these models allows us to assess the robustness and adaptability of current state-of-the-art language models in interpreting and contextualizing new phrases and slang, reflecting real-world language evolution.

\paragraph{Baselines} Baselines comprised direct inquiry (Direct) \cite{NEURIPS2022_b1efde53}, Chain-of-Thought (CoT) \cite{wei2022chain}, CauView \cite{wang-etal-2023-causal}, and our \textsc{Focus} approaches. In each case, the language model output was parsed and compared with the ground truth. We implement the CauView method in our experiment design by using a two-stage prompt inquiry, due to its lack of a direct inquiry step.

%\paragraph{\textsc{Focus}} The outputs from the first three stages were processed by a parser that retained only the final conclusions, filtering out the model's ``reasoning" process. In the fourth stage, the language model was briefed about the concise procedures and objectives of the previous three methods. This briefing included potential key information each method might yield and the risks of information omission. Additionally, to control for syntactic variability and minimize error, the prompts for all baselines, including \textsc{Focus}, incorporated the same three sample entries for few-shot in-context learning. This ensured that the language model's output syntax closely resembled that of the dataset.

\subsection{Results}

\subsection{Experimental Results}

\paragraph{Results on the factual dataset}
Table \ref{table:factual_results} summarizes the findings that the \textsc{Focus} method demonstrates exceptional performance across all models. For GPT-4, \textsc{Focus} achieves the highest $\text{F}_1$ score of 0.4446, precision of 0.4280, recall of 0.4714, and an accuracy of 88.2\%. Similarly, Claude 3 under \textsc{Focus} secures an $\text{F}_1$ score of 0.4596 and an accuracy of 89.7\%, while Mistral 7B records an $\text{F}_1$ score of 0.3703 with an accuracy of 76.0\%. These metrics highlight the effectiveness of the \textsc{Focus} method in enhancing the interpretative capabilities of language models.

\begin{table*}[ht]
    \centering
    \resizebox{2\columnwidth}{!}{
    \begin{tabular}{llcccccccccc}
    \toprule
    \multirow{2}{*}{\textbf{Model}} & \multirow{2}{*}{\textbf{Prompting Method}} & \multirow{2}{*}{$\textbf{F}_\textbf{1}$} & \multirow{2}{*}{\textbf{Precision}} & \multirow{2}{*}{\textbf{Recall}} & \multirow{2}{*}{\textbf{BLUE}} & \multicolumn{3}{c}{\textbf{ROUGE}} & \multirow{2}{*} {\textbf{Similarity}}  & \multirow{2}{*}{\textbf{SimCSE}} &\multirow{2}{*}{\textbf{ACC (\%)}} \\ 
    \cline{7-9} 
    \multicolumn{6}{c}{} &  1 & 2 & \textit{L} & & \\
    % \cline{6-8} \\

    \midrule
    \multirow{4}{*}{\textbf{Mistral-7B}}
    & Direct \cite{NEURIPS2022_b1efde53} & 0.1765 & 0.2402 & 0.1461 & 0.0327 & 0.1983 & 0.0396 & 0.1378 & 0.1520 & 0.5488 & 20.0 \\
    & CoT \cite{wei2022chain} & 0.3318 & 0.3018 & 0.3741 & 0.1849 & 0.3593 & 0.1594 & 0.2734 & 0.3960 & 0.6692 & 56.2 \\
    & CauView \cite{wang-etal-2023-causal} & 0.2922 & 0.2567 & 0.3416 & 0.1431 & 0.3197 & 0.1233 & 0.2365 & 0.3515 & 0.6309 & 46.3 \\
    & \textbf{\textsc{Focus} (Ours)} & \textbf{0.3935} & \textbf{0.3928} & \textbf{0.4005} & \textbf{0.2498} & \textbf{0.4151} & \textbf{0.1891} & \textbf{0.3199} & \textbf{0.5079} & \textbf{0.7343} & \textbf{77.5} \\
    
    \midrule
    \multirow{4}{*}{\textbf{GPT-4}}
    & Direct \cite{NEURIPS2022_b1efde53} & 0.2050 & 0.3018 & 0.1705 & 0.0426 & 0.2341 & 0.0503 & 0.1693 & 0.1645 & 0.5735 & 21.0 \\
    & CoT \cite{wei2022chain} & 0.3821 & 0.3573 & 0.4247 & 0.2189 & 0.4091 & 0.1841 & 0.3147 & 0.4383 & 0.7241 & 62.6 \\
    & CauView \cite{wang-etal-2023-causal} & 0.3357 & 0.3024 & 0.3919 & 0.1744 & 0.3701 & 0.1467 & 0.2812 & 0.3815 & 0.6715 & 47.2 \\
    
    & \textbf{\textsc{Focus} (Ours)} & \textbf{0.4532} & \textbf{0.4598} & \textbf{0.4551} & \textbf{0.3017} & \textbf{0.4763} & \textbf{0.2273} & \textbf{0.3722} & \textbf{0.5645} & \textbf{0.8065} & \textbf{84.9} \\

    \midrule
    \multirow{4}{*}{\textbf{Claude 3}}
    & Direct \cite{NEURIPS2022_b1efde53} & 0.2123 & 0.3143 & 0.1779 & 0.0454 & 0.2435 & 0.0496 & 0.1769 & 0.1733 & 0.5909 & 23.1 \\
    & CoT \cite{wei2022chain} & 0.3928 & 0.3605 & 0.4405 & 0.2143 & 0.4138 & 0.1867 & 0.3184 & 0.4499 & 0.7398 & 64.7 \\
    & CauView \cite{wang-etal-2023-causal} & 0.3468 & 0.3101 & 0.4062 & 0.1662 & 0.3791 & 0.1490 & 0.2879 & 0.4012 & 0.6941 & 52.4 \\
    & \textbf{\textsc{Focus} (Ours)} & \textbf{0.4636} & \textbf{0.4739} & \textbf{0.4618} & \textbf{0.3132} & \textbf{0.4894} & \textbf{0.2361} & \textbf{0.3867} & \textbf{0.5783} & \textbf{0.8216} & \textbf{86.8} \\

    \bottomrule
    \end{tabular}
    }
    \caption{Performance results on the counterfactual dataset. For results on more models, see Appendix \ref{appendix:new_counterfactual_results}.}
    \label{table:counterfactual_results}

\end{table*}

\begin{table*}[ht]
    \centering
    \resizebox{2\columnwidth}{!}{
    \begin{tabular}{lcccccccccc}
    \toprule
    \textbf{Experiment} & $\textbf{F}_\textbf{1}$ & \textbf{Precision} & \textbf{Recall} & \textbf{BLUE} & \textbf{ROUGE-1} & \textbf{ROUGE-2} & \textbf{ROUGE-L} & \textbf{Similarity} & \textbf{SimCSE} & \textbf{ACC (\%)} \\ 
    \midrule
    w/o MEA & 0.4366 & 0.4380 & 0.4484 & 0.2766 & 0.4586 & 0.2059 & 0.3556 & 0.8821 & 0.8014 & 82.0 \\
    w/o ERI & 0.4283 & 0.4300 & 0.4371 & 0.2814 & 0.4593 & 0.2117 & 0.3547 & 0.9021 & 0.8092 & 84.0 \\
    \bottomrule
    \end{tabular}
    }
    \caption{Results of ablation experiments on GPT-4. For results on more models, see Appendix \ref{appendix:new_ablation_study}.}
    \label{table:ablation_study}
    
\end{table*}

\paragraph{Results on the counterfactual dataset}
Following the analysis of the factual dataset, we extended our evaluation to the counterfactual dataset, which presents hypothetical scenarios altering real-world language usage.
As detailed in Table \ref{table:counterfactual_results}, the \textsc{Focus} method outshines other techniques, particularly with GPT-4, achieving an $\text{F}_1$ score of 0.4532, precision of 0.4598, recall of 0.4551, and an accuracy of 84.9\%. Claude 3 also performs well under \textsc{Focus}, securing an $\text{F}_1$ score of 0.4636 and an accuracy of 86.8\%. Mistral 7B shows a solid performance, with an $\text{F}_1$ score of 0.3935 and an accuracy of 77.5\%. These results underscore \textsc{Focus}'s robust ability to navigate the challenges posed by modified linguistic contexts.

\subsection{Ablation Study}
\label{ablation_study}

Our ablation study, focusing on the \textsc{Focus} methodology's components MEI and ERI on the factual dataset. As shown in Table \ref{table:ablation_study}, these components significantly enhance the interpretative capabilities of the model. Extended ablation results are provided in Appendix~\ref{appendix:new_ablation_study}.
\paragraph{MEI} The MEI stage, represented mathematically as \( P(Y|W,X,E;\mathcal{M}) \), where \( Y \) is the interpretation, \( W \) the phrase, \( X \) the context, and \( E \) the entities, critically influences the model's performance. Its exclusion (w/o MEI) reduced the $\text{F}_1$ score to 0.4366 from 0.4446. This result illustrates MEA's role in disentangling the direct influence of \( W \) and \( X \) from confounding entities \( E \), vital for context-driven interpretation.
\paragraph{ERI} The ERI stage, which examines the causal links \( X \rightarrow Y \leftarrow E \) and \( X \rightarrow E \), also shows significant impact. Removing ERI (w/o ERI) decreased the $\text{F}_1$ score to 0.4283. ERI's function in the model, isolating the entity's influence and exploring alternative causal pathways, proves essential for nuanced language interpretation.

While incorporating either MEI or ERI individually into direct inquiry enhances model performance, their combined use in the \textsc{Focus} framework is indispensable for achieving optimal results. This synergy underscores the importance of a comprehensive causal analysis, balancing context and entity dynamics, for the nuanced interpretation of evolving linguistic phenomena.

\section{Related Work}

\subsection{Knowledge Update Methods}
LLMs enhance knowledge through parameter-efficient fine-tuning methods like task-specific parameter addition \cite{houlsby2019parameter} and low-rank adaptation (LoRA) \cite{lora2021less, pfeiffer2020adapterhub}. However, these methods face challenges such as computational demands, catastrophic forgetting, and reduced task-specific effectiveness \cite{lester2021power}. BitFit \cite{ben2021bitfit} simplifies fine-tuning but relies heavily on dataset quality. Trade-offs in computational demands, flexibility, and task compatibility are essential considerations. Retrieval-augmented generation \cite{lewis2020retrieval} and in-context learning-based knowledge editing \cite{zhong-etal-2023-mquake} offer dynamic integration of external information, focusing more on fact retrieval than on enhancing deeper understanding.

\subsection{Entity Bias and Shortcuts in LLMs}
Entity bias \cite{peng-etal-2020-learning, longpre-etal-2021-entity, wang-etal-2022-rely, wang-etal-2023-causal} and shortcuts \cite{du-etal-2021-towards, PrOntoQA} in LLMs lead to oversimplified language processing, relying on specific entities or dataset-driven patterns. Entity bias skews model predictions towards certain entities, while shortcuts encompass simplified heuristics, focusing on identifiable features or aspects of the input \cite{du2022shortcut}. These patterns limit the models' understanding and generation of nuanced language, affecting generalization and robustness.

\subsection{Causal Intervention Solutions}
Causal interventions for debiasing and mitigating shortcuts have gained prominence. \cite{tian2022debiasing} and \cite{zhou-etal-2023-causal} focus on causal inference and invariant learning for debiasing. CausaLM \cite{feder2021causalm} provides causal-based model explanations, addressing previous tools' limitations. Counterfactual methods for debiasing \cite{chen-etal-2023-causal} and eliminating shortcuts \cite{wen-etal-2022-autocad} have shown promise, though limitations remain in targeting white-box models and retraining requirements. 
%Our research addresses these limitations through contextual learning and causal inference for bias and shortcut elimination without retraining.

\section{Conclusion}

In this work, we have explored the dynamic and evolving nature of internet language, particularly slang and memes, and their impact on the adaptability of LLMs. Our study introduced a novel benchmark, \textsc{Slang}, to assess LLMs' proficiency in comprehending emerging linguistic trends. Additionally, we proposed the \textsc{Focus} methodology, which utilizes causal inference to enhance understanding of new concepts, going beyond other methods in terms of precision and relevance. Our approach involves the construction of datasets from UrbanDictionary, a platform known for user-generated content that reflects current language trends. We incorporated both factual and counterfactual instances to provide diverse linguistic contexts. Factual instances are drawn directly from the UrbanDictionary entries, while counterfactual instances are created by altering real-world examples to assess the models' adaptability to hypothetical scenarios. The results from our experiments demonstrate the enhanced capability of LLMs, equipped with our \textsc{Focus} method, to adapt to the rapid evolution of online language. This research contributes to the field of natural language processing by emphasizing the importance of contextual understanding and adaptability in LLMs. Our findings suggest that LLMs can effectively navigate the complexities of evolving human communication when equipped with robust methodologies like \textsc{Focus} and evaluated against benchmarks such as \textsc{Slang}.

\section*{Limitations}
The main limitation of this work is that it does not fully address the complexities of linguistic evolution in non-English or morphologically rich languages. Therefore, future work should explore a wider range of linguistic scenarios and extend our methodology to other languages and linguistic contexts. Additionally, the \textsc{Focus} methodology, despite its effectiveness in enhancing the understanding of emerging linguistic phenomena, has a higher computational complexity compared to some traditional approaches. This might not only increase the computational demands but also introduce delays when deployed on mobile devices, which could hinder real-time applications. Such issues necessitate further optimization to reduce computational load and improve efficiency for mobile and other constrained environments. Moreover, applying this methodology to downstream tasks might encounter challenges related to data processing, as the sources of new concepts may not be readily accessible. This could require the use of external tools to gather relevant data, potentially making the data collection process time-consuming and variable depending on the specific task.
%Although the \textsc{Focus} methodology has demonstrated its effectiveness in enhancing the understanding of emerging linguistic phenomena, it is worth noting that its computational complexity may be higher compared to some traditional approaches. Additionally, our approach, although innovative, does not fully address the complexities of linguistic evolution in non-English or morphologically rich languages. Future work should explore more diverse datasets and extend our methodology to other languages and linguistic contexts.

\section*{Ethics Statement}
Our research acknowledges that while methods like the \textsc{Slang} benchmark and \textsc{Focus} approach enhance LLM's understanding of Internet language, they cannot entirely eliminate the propagation of harmful content. Users must exercise caution and cultural sensitivity, especially when interpreting slang and memes, to avoid reinforcing stereotypes or biases. Our work encourages responsible use, emphasizing the importance of respecting diverse linguistic origins and the natural evolution of language.

\section*{Acknowledge}
This paper is partially supported by the National Science Foundation of China under Grant No.U21B2046 and 6237075198, and National Key R\&D Program of China (No.2023YFC3305303).

\bibliography{anthology,custom}

% \appendix

% \section{Transformation Template}
% \label{sec:appendix}

% This is an appendix.

% \section*{Appendix}
\label{sec:appendix}

\appendix

\section{Detail Experimental Setup}\label{appendix:setup}
In this section, we provide a comprehensive overview of our experimental setup, including the models used, API parameters, preprocessing steps, and the standardization process for definitions. This setup ensures that our evaluations are thorough, replicable, and provide meaningful insights into the performance of various large language models (LLMs).

\subsection{Models}
We evaluated both closed-source and open-source LLMs to ensure a broad assessment.

\paragraph{Closed-source models.} 
\begin{itemize}
    \item GPT-3.5: \\ \texttt{gpt-3.5-turbo-1106} 
    \item GPT-4:\\  \texttt{gpt-4-0613}
    \item Claude 3 Opus:  \\ \texttt{claude-3-opus-20240229}
    \item Claude 3 Sonnet: \\ \texttt{claude-3-sonnet-20240229}
    \item Claude 3 Haiku: \\ \texttt{claude-3-haiku-20240307}
\end{itemize}

\paragraph{Open-source models.} 
\begin{itemize}
    \item Mistral 7B~\cite{jiang2023mistral} : \\ \texttt{mistral-7b-instruct-v0.1}
    \item LLaMA 7B: \\ \texttt{llama-2-7b-chat}
\end{itemize}

\subsection{API Parameters}
To ensure consistency and comparability in our evaluations, we used the following hyperparameters when calling the LLM API:
\begin{itemize}
    \item \textbf{Temperature:} 0.7
    \item \textbf{Max tokens:} 512
\end{itemize}

\subsection{Preprocessing}
The selection of 408 samples from 7220 was based strictly on the data filtering strategy described in Section \ref{subsec:dataset} of our paper. Although this number of test samples may seem small for an unfiltered dataset, it is substantial compared to existing LLM benchmarks:
\begin{itemize}
    \item MaliciousInstruct~\citep{huang2023catastrophic}:\\ \textbf{100} samples
    \item HumanEval~\citep{chen2021evaluating}:\\ \textbf{163} samples
    \item AdvBench~\citep{zou2023universal}:\\ \textbf{500} samples
    \item HarmBench~\citep{mazeika2024harmbench}:\\ \textbf{400} unimodal samples
\end{itemize}
Additionally, we ensured the quality and diversity of the samples, as illustrated in Figure 6. Our chosen test samples were novel to the LLM, guaranteeing a comprehensive evaluation of model performance by maintaining the integrity and representativeness of the dataset.

\subsection{Costs}
The experiments conducted using closed-source models incurred a total cost of approximately \$500 (includes API discounts). For the open-source models, the experiments were run on a server with 4 NVIDIA Tesla A100 GPUs for a duration of 14 hours.

This breakdown of costs highlights the computational and financial resources required to conduct comprehensive evaluations of large language models. The use of both closed-source and open-source models ensures a diverse and robust assessment, while the detailed cost analysis provides transparency regarding the experimental setup.

\section{Explanation Standardization}\label{sec:standardization}
Since the definitions were user-generated, they varied widely in language style and structural format. To ensure uniformity, we standardized these explanations using a consistent template.

We used the following template to standardize the explanations in the dataset:
\begin{quote}
[P] \textbf{refers to} [B]. \textbf{It is often used} [C]. \textbf{This expression} [A].
\end{quote}

\begin{itemize}
    \item \textbf{P:} Phrase
    \item \textbf{B:} Basic description of the word
    \item \textbf{C:} Context or situation of usage
    \item \textbf{A:} Additional details like connotations, emotions, or typical reactions associated with the word
\end{itemize}

For example, the original explanation of the phrase \textit{The Winter Arc} was: 
\begin{quote}
``A time where every man must face the mental and physical challenges of winter. A time to put your head down and get things done''
\end{quote}
After standardization, it became:
\begin{quote}
``The Winter Arc \textbf{refers to} a time when people deal with the cold and hard parts of winter. \textbf{It is often used} to talk about staying strong and getting work done even when it's cold and challenging outside. \textbf{This expression} suggests that people are being tough and focused.''
\end{quote}

\section{Additional Experimental Results}
\label{appendix:additional_results}

We provide a comprehensive summary of the extended evaluation conducted on various language models including GPT-3.5, GPT-4, several versions of Claude 3, Mistral 7B, and LLaMA 2-7B \cite{touvron2023llama}. Each model undergoes assessment using a range of prompting methods, such as Direct, CoT, CauView, and our \textsc{Focus} method.

\subsection{Factual Dataset}
\label{appendix:new_factual_results}
In the factual dataset, the \textsc{Focus} prompting method propels GPT-4 and Claude 3 Opus models to the highest performance, with GPT-4 achieving an $\text{F}_1$ score of 0.4446, precision of 0.4280, recall of 0.4714, and accuracy of 88.2\%. Claude 3 Opus closely follows with an $\text{F}_1$ score of 0.4596 and accuracy of 89.7\%. The LLaMA 2-7B also exhibits commendable improvements, confirming the efficacy of \textsc{Focus} across diverse architectures. These results are shown in Table \ref{table:new_factual_results}.

\begin{table*}[ht]
\centering
\resizebox{2\columnwidth}{!}{
\begin{tabular}{llcccccccccc}
\toprule
\multirow{2}{*}{\textbf{Model}} & \multirow{2}{*}{\textbf{Prompting Method}} & \multirow{2}{*}{$\textbf{F}_\textbf{1}$} & \multirow{2}{*}{\textbf{Precision}} & \multirow{2}{*}{\textbf{Recall}} & \multirow{2}{*}{\textbf{BLEU}} & \multicolumn{3}{c}{\textbf{ROUGE}} & \multirow{2}{*} {\textbf{Similarity}} & \multirow{2}{*}{\textbf{SimCSE}} &\multirow{2}{*}{\textbf{ACC (\%)}} \\
\cline{7-9}
\multicolumn{6}{c}{} & 1 & 2 & \textit{L} & & \\
\midrule
\multirow{4}{*}{\textbf{GPT-3.5}}
& Direct & 0.2219 & 0.1541 & 0.4320 & 0.0477 & 0.2410 & 0.0547 & 0.1531 & 0.1943 & 0.6806 & 47.6 \\
& CoT & 0.3922 & 0.3585 & 0.4583 & 0.2273 & 0.4181 & 0.1822 & 0.3135 & 0.5292 & 0.7653 & 76.4 \\
& CauView & 0.3500 & 0.2487 & \textbf{0.4885} & 0.1676 & 0.3777 & 0.1506 & 0.2676 & 0.5612 & 0.7868 & 72.0 \\
& \textbf{\textsc{Focus} (Ours)} & \textbf{0.4292} & \textbf{0.4153} & 0.4524 & \textbf{0.2798} & \textbf{0.4541} & \textbf{0.2131} & \textbf{0.3481} & \textbf{0.5748} & \textbf{0.8017} & \textbf{84.5} \\
\midrule
\multirow{4}{*}{\textbf{GPT-4}}
& Direct & 0.2308 & 0.3474 & 0.1917 & 0.0483 & 0.2597 & 0.0606 & 0.1859 & 0.1616 & 0.6476 & 47.2 \\
& CoT & 0.4123 & 0.3947 & 0.4244 & 0.2384 & 0.4370 & 0.1927 & 0.3299 & 0.5521 & 0.7883 & 79.3 \\
& CauView & 0.3602 & 0.3444 & 0.3948 & 0.1987 & 0.3917 & 0.1643 & 0.3032 & 0.5515 & 0.7636 & 74.7 \\
& \textbf{\textsc{Focus} (Ours)} & \textbf{0.4446} & \textbf{0.4280} & \textbf{0.4714} & \textbf{0.3177} & \textbf{0.4721} & \textbf{0.2332} & \textbf{0.3652} & \textbf{0.6032} & \textbf{0.8216} & \textbf{88.2} \\
\midrule
\multirow{4}{*}{\textbf{Claude 3 Opus}}
& Direct & 0.2395 & 0.3538 & 0.2171 & 0.0552 & 0.2645 & 0.0673 & 0.1904 & 0.1947 & 0.6714 & 51.4 \\
& CoT & 0.4276 & 0.4082 & 0.4606 & 0.2471 & 0.4492 & 0.2014 & 0.3371 & 0.5628 & 0.7948 & 81.2 \\
& CauView & 0.3752 & 0.3550 & 0.4229 & 0.2015 & 0.4035 & 0.1707 & 0.3041 & 0.5778 & 0.7932 & 76.6 \\
& \textbf{\textsc{Focus} (Ours)} & \textbf{0.4596} & \textbf{0.4452} & \textbf{0.4827} & \textbf{0.3264} & \textbf{0.4835} & \textbf{0.2373} & \textbf{0.3729} & \textbf{0.6109} & \textbf{0.8354} & \textbf{89.7} \\
\midrule
\multirow{4}{*}{\textbf{Claude 3 Sonnet}}
& Direct & 0.2251 & 0.3411 & 0.2015 & 0.0495 & 0.2541 & 0.0592 & 0.1813 & 0.1587 & 0.6417 & 46.8 \\
& CoT & 0.4075 & 0.3895 & 0.4202 & 0.2353 & 0.4312 & 0.1892 & 0.3247 & 0.5467 & 0.7834 & 78.7 \\
& CauView & 0.3562 & 0.3393 & 0.3897 & 0.1946 & 0.3872 & 0.1602 & 0.2951 & 0.5462 & 0.7586 & 73.9 \\
& \textbf{\textsc{Focus} (Ours)} & \textbf{0.4391} & \textbf{0.4238} & \textbf{0.4653} & \textbf{0.3128} & \textbf{0.4676} & \textbf{0.2293} & \textbf{0.3601} & \textbf{0.5982} & \textbf{0.8167} & \textbf{87.6} \\
\midrule
\multirow{4}{*}{\textbf{Claude 3 Haiku}}
& Direct & 0.2137 & 0.3105 & 0.1912 & 0.0463 & 0.2356 & 0.0558 & 0.1652 & 0.1849 & 0.6639 & 48.7 \\
& CoT & 0.3853 & 0.3530 & 0.4271 & 0.2196 & 0.4085 & 0.1776 & 0.3050 & 0.5217 & 0.7559 & 75.2 \\
& CauView & 0.3384 & 0.3032 & 0.3816 & 0.1782 & 0.3663 & 0.1498 & 0.2743 & 0.5394 & 0.7562 & 70.6 \\
& \textbf{\textsc{Focus} (Ours)} & \textbf{0.4086} & \textbf{0.3875} & \textbf{0.4308} & \textbf{0.2759} & \textbf{0.4326} & \textbf{0.2041} & \textbf{0.3303} & \textbf{0.5648} & \textbf{0.7896} & \textbf{82.8} \\
\midrule
\multirow{4}{*}{\textbf{Mistral-7B}}
& Direct & 0.1869 & 0.2655 & 0.1684 & 0.0372 & 0.2078 & 0.0481 & 0.1452 & 0.1507 & 0.6084 & 43.1 \\
& CoT & 0.3453 & 0.3227 & 0.3706 & 0.1978 & 0.3657 & 0.1602 & 0.2744 & 0.4713 & 0.7132 & 68.5 \\
& CauView & 0.3012 & 0.2787 & 0.3384 & 0.1561 & 0.3272 & 0.1336 & 0.2449 & 0.4862 & 0.7127 & 64.3 \\
& \textbf{\textsc{Focus} (Ours)} & \textbf{0.3703} & \textbf{0.3555} & \textbf{0.3878} & \textbf{0.2493} & \textbf{0.3942} & \textbf{0.1866} & \textbf{0.3011} & \textbf{0.5121} & \textbf{0.7469} & \textbf{76.0} \\
\midrule
\multirow{4}{*}{\textbf{LLaMA 2-7B}}
& Direct & 0.1581 & 0.2115 & 0.1361 & 0.0282 & 0.1795 & 0.0378 & 0.1254 & 0.1134 & 0.5792 & 40.5 \\
& CoT & 0.3174 & 0.2884 & 0.3491 & 0.1746 & 0.3344 & 0.1407 & 0.2515 & 0.4628 & 0.6844 & 64.4 \\
& CauView & 0.2717 & 0.2441 & 0.3109 & 0.1416 & 0.2951 & 0.1152 & 0.2242 & 0.4669 & 0.6718 & 59.5 \\
& \textbf{\textsc{Focus} (Ours)} & \textbf{0.3391} & \textbf{0.3282} & \textbf{0.3549} & \textbf{0.2043} & \textbf{0.3619} & \textbf{0.1579} & \textbf{0.2771} & \textbf{0.4873} & \textbf{0.7117} & \textbf{70.8} \\
% \midrule
% \multirow{4}{*}{\textbf{OLMo-7B}}
% & Direct & 0.1650 & 0.2317 & 0.1453 & 0.0317 & 0.1874 & 0.0405 & 0.1311 & 0.1221 & 0.5911 & 41.8 \\
% & CoT & 0.3293 & 0.2994 & 0.3621 & 0.1815 & 0.3442 & 0.1454 & 0.2586 & 0.4793 & 0.7004 & 66.4 \\
% & CauView & 0.2830 & 0.2551 & 0.3239 & 0.1483 & 0.3076 & 0.1199 & 0.2314 & 0.4839 & 0.6896 & 61.6 \\
% & \textbf{\textsc{Focus} (Ours)} & \textbf{0.3538} & \textbf{0.3425} & \textbf{0.3704} & \textbf{0.2155} & \textbf{0.3778} & \textbf{0.1659} & \textbf{0.2889} & \textbf{0.5026} & \textbf{0.7297} & \textbf{73.1} \\
\bottomrule
\end{tabular}
}
\caption{Performance results on the factual dataset.}
\label{table:new_factual_results}
\end{table*}

\subsection{Counterfactual Dataset}
\label{appendix:new_counterfactual_results}
When analyzing the counterfactual dataset, which involves challenges from hypothetical language alterations, \textsc{Focus} maintains the lead under GPT-4, achieving an $\text{F}_1$ score of 0.4532 and accuracy of 84.9\%. Claude 3 Opus exhibits robustness in this modified context with an $\text{F}_1$ score of 0.4636 and accuracy of 86.8\%. The lower-resource models such as LLaMA 2-7B also displays significant gains, demonstrating the adaptability of \textsc{Focus} to a wide range of models and scenarios. These results are shown in Table \ref{table:new_counterfactual_results}.

\begin{table*}[ht]
\centering
\resizebox{2\columnwidth}{!}{
\begin{tabular}{llcccccccccc}
\toprule
\multirow{2}{*}{\textbf{Model}} & \multirow{2}{*}{\textbf{Prompting Method}} & \multirow{2}{*}{$\textbf{F}_\textbf{1}$} & \multirow{2}{*}{\textbf{Precision}} & \multirow{2}{*}{\textbf{Recall}} & \multirow{2}{*}{\textbf{BLEU}} & \multicolumn{3}{c}{\textbf{ROUGE}} & \multirow{2}{*} {\textbf{Similarity}} & \multirow{2}{*}{\textbf{SimCSE}} &\multirow{2}{*}{\textbf{ACC (\%)}} \\
\cline{7-9}
\multicolumn{6}{c}{} & 1 & 2 & \textit{L} & & \\
\midrule
\multirow{4}{*}{\textbf{GPT-3.5}}
& Direct & 0.1922 & 0.1252 & \textbf{0.4497} & 0.0371 & 0.2120 & 0.0477 & 0.1328 & 0.1936 & 0.6051 & 24.3 \\
& CoT & 0.3576 & 0.3138 & 0.4407 & 0.2098 & 0.3865 & 0.1713 & 0.2932 & 0.4390 & 0.7232 & 62.3 \\
& CauView & 0.3161 & 0.2489 & 0.4471 & 0.1464 & 0.3439 & 0.1297 & 0.2491 & 0.4031 & 0.6954 & 54.6 \\
& \textbf{\textsc{Focus} (Ours)} & \textbf{0.4078} & \textbf{0.4018} & 0.4294 & \textbf{0.2519} & \textbf{0.4344} & \textbf{0.1955} & \textbf{0.3339} & \textbf{0.5594} & \textbf{0.7836} & \textbf{83.4} \\
\midrule
\multirow{4}{*}{\textbf{GPT-4}}
& Direct & 0.2050 & 0.3018 & 0.1705 & 0.0426 & 0.2341 & 0.0503 & 0.1693 & 0.1645 & 0.5735 & 21.0 \\
& CoT & 0.3821 & 0.3573 & 0.4247 & 0.2189 & 0.4091 & 0.1841 & 0.3147 & 0.4383 & 0.7241 & 62.6 \\
& CauView & 0.3357 & 0.3024 & 0.3919 & 0.1744 & 0.3701 & 0.1467 & 0.2812 & 0.3815 & 0.6715 & 47.2 \\
& \textbf{\textsc{Focus} (Ours)} & \textbf{0.4532} & \textbf{0.4598} & \textbf{0.4551} & \textbf{0.3017} & \textbf{0.4763} & \textbf{0.2273} & \textbf{0.3722} & \textbf{0.5645} & \textbf{0.8065} & \textbf{84.9} \\
\midrule
\multirow{4}{*}{\textbf{Claude 3 Opus}}
& Direct & 0.2123 & 0.3143 & 0.1779 & 0.0454 & 0.2435 & 0.0496 & 0.1769 & 0.1733 & 0.5909 & 23.1 \\
& CoT & 0.3928 & 0.3605 & 0.4405 & 0.2143 & 0.4138 & 0.1867 & 0.3184 & 0.4499 & 0.7398 & 64.7 \\
& CauView & 0.3468 & 0.3101 & 0.4062 & 0.1662 & 0.3791 & 0.1490 & 0.2879 & 0.4012 & 0.6941 & 52.4 \\
& \textbf{\textsc{Focus} (Ours)} & \textbf{0.4636} & \textbf{0.4739} & \textbf{0.4618} & \textbf{0.3132} & \textbf{0.4894} & \textbf{0.2361} & \textbf{0.3867} & \textbf{0.5783} & \textbf{0.8216} & \textbf{86.8} \\
\midrule
\multirow{4}{*}{\textbf{Claude 3 Sonnet}}
& Direct & 0.1984 & 0.2886 & 0.1637 & 0.0398 & 0.2261 & 0.0445 & 0.1628 & 0.1539 & 0.5540 & 20.5 \\
& CoT & 0.3752 & 0.3439 & 0.4211 & 0.2038 & 0.3972 & 0.1790 & 0.3053 & 0.4249 & 0.7080 & 61.1 \\
& CauView & 0.3302 & 0.2952 & 0.3873 & 0.1578 & 0.3617 & 0.1421 & 0.2745 & 0.3776 & 0.6596 & 49.6 \\
& \textbf{\textsc{Focus} (Ours)} & \textbf{0.4416} & \textbf{0.4521} & \textbf{0.4398} & \textbf{0.2956} & \textbf{0.4666} & \textbf{0.2249} & \textbf{0.3682} & \textbf{0.5502} & \textbf{0.7874} & \textbf{83.2} \\
\midrule
\multirow{4}{*}{\textbf{Claude 3 Haiku}}
& Direct & 0.1853 & 0.2392 & 0.1482 & 0.0334 & 0.2018 & 0.0384 & 0.1405 & 0.1708 & 0.6026 & 22.7 \\
& CoT & 0.3458 & 0.3044 & 0.3898 & 0.1923 & 0.3649 & 0.1589 & 0.2763 & 0.4297 & 0.7196 & 60.5 \\
& CauView & 0.3038 & 0.2583 & 0.3540 & 0.1490 & 0.3280 & 0.1234 & 0.2431 & 0.3860 & 0.6786 & 51.7 \\
& \textbf{\textsc{Focus} (Ours)} & \textbf{0.3959} & \textbf{0.3885} & \textbf{0.4086} & \textbf{0.2462} & \textbf{0.4164} & \textbf{0.1868} & \textbf{0.3180} & \textbf{0.5428} & \textbf{0.7716} & \textbf{81.3} \\
\midrule
\multirow{4}{*}{\textbf{Mistral 7B}}
& Direct & 0.1765 & 0.2402 & 0.1461 & 0.0327 & 0.1983 & 0.0396 & 0.1378 & 0.1520 & 0.5488 & 20.0 \\
& CoT & 0.3318 & 0.3018 & 0.3741 & 0.1849 & 0.3593 & 0.1594 & 0.2734 & 0.3960 & 0.6692 & 56.2 \\
& CauView & 0.2922 & 0.2567 & 0.3416 & 0.1431 & 0.3197 & 0.1233 & 0.2365 & 0.3515 & 0.6309 & 46.3 \\
& \textbf{\textsc{Focus} (Ours)} & \textbf{0.3935} & \textbf{0.3928} & \textbf{0.4005} & \textbf{0.2498} & \textbf{0.4151} & \textbf{0.1891} & \textbf{0.3199} & \textbf{0.5079} & \textbf{0.7343} & \textbf{77.5} \\
\midrule
\multirow{4}{*}{\textbf{LLaMA 2-7B}}
& Direct & 0.1421 & 0.1680 & 0.1107 & 0.0207 & 0.1495 & 0.0281 & 0.1021 & 0.1033 & 0.4606 & 13.7 \\
& CoT & 0.2619 & 0.2381 & 0.2950 & 0.1472 & 0.2837 & 0.1255 & 0.2159 & 0.3151 & 0.5475 & 44.5 \\
& CauView & 0.2309 & 0.2026 & 0.2698 & 0.1122 & 0.2520 & 0.0971 & 0.1864 & 0.2762 & 0.5177 & 36.1 \\
& \textbf{\textsc{Focus} (Ours)} & \textbf{0.3087} & \textbf{0.3081} & \textbf{0.3142} & \textbf{0.1954} & \textbf{0.3255} & \textbf{0.1479} & \textbf{0.2510} & \textbf{0.3985} & \textbf{0.6073} & \textbf{60.6} \\
% \midrule
% \multirow{4}{*}{\textbf{OLMo-7B}}
% & Direct & 0.1497 & 0.1867 & 0.1206 & 0.0235 & 0.1599 & 0.0317 & 0.1105 & 0.1119 & 0.4803 & 15.2 \\
% & CoT & 0.2782 & 0.2528 & 0.3132 & 0.1561 & 0.3011 & 0.1332 & 0.2291 & 0.3343 & 0.5779 & 47.2 \\
% & CauView & 0.2453 & 0.2154 & 0.2868 & 0.1191 & 0.2683 & 0.1034 & 0.1981 & 0.2938 & 0.5451 & 38.4 \\
% & \textbf{\textsc{Focus} (Ours)} & \textbf{0.3273} & \textbf{0.3268} & \textbf{0.3332} & \textbf{0.2076} & \textbf{0.3463} & \textbf{0.1575} & \textbf{0.2664} & \textbf{0.4227} & \textbf{0.6390} & \textbf{64.3} \\
\bottomrule
\end{tabular}
}
\caption{Performance results on the counterfactual dataset.}
\label{table:new_counterfactual_results}
\end{table*}

\subsection{Ablation Study}
\label{appendix:new_ablation_study}
We extend the ablation study to more models, focusing on the effect of the MEI and ERI components of \textsc{Focus} on the factual dataset. As shown in Table \ref{table:new_ablation_study}, removing these components leads to a significant performance drop across all models tested. These results further confirm the findings in Section \ref{ablation_study}, demonstrating that MEI and ERI are crucial for achieving optimal performance within \textsc{Focus}.

\subsection{Discussion}
Across both datasets, \textsc{Focus} consistently excels, enabling models to understand and interpret language effectively. This effectiveness is particularly evident in models with diverse capacities and structures, signifying the versatility of \textsc{Focus} in its application to natural language processing tasks. Furthermore, the lower recall score observed in GPT-3.5 is attributed to the model's tendency to provide shorter answers to questions, a characteristic preference of GPT-3.5 itself rather than a limitation of \textsc{Focus}. Although this preference for brevity contributes to a lower recall score, it does not diminish the overall effectiveness of \textsc{Focus}.

% \begin{table*}[ht]
% \centering
% \resizebox{2\columnwidth}{!}{
% \begin{tabular}{llcccccccccc}
% \hline\hline
% \toprule
% \multirow{2}{*}{\textbf{Model}} & \multirow{2}{*}{\textbf{Experiment}} & \multirow{2}{*}{$\textbf{F}_\textbf{1}$} & \multirow{2}{*}{\textbf{Precision}} & \multirow{2}{*}{\textbf{Recall}} & \multirow{2}{*}{\textbf{BLEU}} & \multicolumn{3}{c}{\textbf{ROUGE}} & \multirow{2}{*} {\textbf{Similarity}} & \multirow{2}{*}{\textbf{SimCSE}} &\multirow{2}{*}{\textbf{ACC (\%)}} \
% \cline{7-9}
% \multicolumn{6}{c}{} & 1 & 2 & \textit{L} & & \
% \midrule

% \multirow{2}{*}{\textbf{Mistral 7B}} & w/o MEA & 0.3624 & 0.3477 & 0.3792 & 0.2414 & 0.3864 & 0.1787 & 0.2933 & 0.5032 & 0.7379 & 74.7 \
%  & w/o ERI & 0.3663 & 0.3515 & 0.3833 & 0.2453 & 0.3903 & 0.1816 & 0.2972 & 0.5081 & 0.7421 & 75.4 \
% \midrule
% \multirow{2}{*}{\textbf{Claude 3 Opus}} & w/o MEA & 0.4325 & 0.4160 & 0.4561 & 0.2962 & 0.4565 & 0.2153 & 0.3480 & 0.5806 & 0.8018 & 85.1 \
%  & w/o ERI & 0.4369 & 0.4202 & 0.4608 & 0.3003 & 0.4606 & 0.2184 & 0.3519 & 0.5857 & 0.8062 & 85.8 \
% \midrule

% \bottomrule
% \end{tabular}
% }
% \caption{Results of ablation experiments on Mistral, Claude 3 Opus, and OLMo.}
% \label{table:new_ablation_study}
% \end{table*}

\begin{table*}[ht]
    \centering
    \resizebox{2\columnwidth}{!}{ % Ensure this scales as expected
    \begin{tabular}{lccccccccccc}
    \toprule
    \textbf{Model} & \textbf{Experiment} & $\mathbf{F}_1$ & \textbf{Precision} & \textbf{Recall} & \textbf{BLEU} & \textbf{ROUGE-1} & \textbf{ROUGE-2} & \textbf{ROUGE-L} & \textbf{Similarity} & \textbf{SimCSE} & \textbf{ACC(\%)} \\
    \midrule
    \multirow{2}{*}{\textbf{Claude 3 Opus}} & w/o MEA & 0.4325 & 0.4160 & 0.4561 & 0.2962 & 0.4565 & 0.2153 & 0.3480 & 0.5806 & 0.8018 & 85.1 \\
    & w/o ERI & 0.4369 & 0.4202 & 0.4608 & 0.3003 & 0.4606 & 0.2184 & 0.3519 & 0.5857 & 0.8062 & 85.8 \\
    \midrule
    \multirow{2}{*}{\textbf{GPT-4}} & MEA & 0.4366 & 0.4380 & 0.4484 & 0.2766 & 0.4586 & 0.2059 & 0.3556 & 0.8821 & 0.8014 & 82.0 \\
    &  w/o ERI & 0.4283 & 0.4300 & 0.4371 & 0.2814 & 0.4593 & 0.2117 & 0.3547 & 0.9021 & 0.8092 & 84.0 \\
    \midrule
    \multirow{2}{*}{\textbf{Mistral 7B}} & w/o MEA & 0.3624 & 0.3477 & 0.3792 & 0.2414 & 0.3864 & 0.1787 & 0.2933 & 0.5032 & 0.7379 & 74.7 \\
    &  w/o ERI & 0.3663 & 0.3515 & 0.3833 & 0.2453 & 0.3903 & 0.1816 & 0.2972 & 0.5081 & 0.7421 & 75.4 \\

    \bottomrule
    \end{tabular}
    }
    \caption{Results of ablation experiments on more models.}
    \label{table:new_ablation_study}
\end{table*}

% \begin{table*}[ht]
%     \centering
%     \resizebox{2\columnwidth}{!}{
%     \begin{tabular}{lccccccccccc}
%     \hline\hline
%     \toprule
%     \textbf{Model} & \textbf{Experiment} & $\textbf{F}_\textbf{1}$ & \textbf{Precision} & \textbf{Recall} & \textbf{BLUE} & \textbf{ROUGE-1} & \textbf{ROUGE-2} & \textbf{ROUGE-L} & \textbf{Similarity} & \textbf{SimCSE} & \textbf{ACC (\%)} \ 
%     \midrule
%     \multirow{2}{*}{\textbf{Mistral 7B}} & w/o MEA & 0.4366 & 0.4380 & 0.4484 & 0.2766 & 0.4586 & 0.2059 & 0.3556 & 0.8821 & 0.8014 & 82.0 \
%     & w/o ERI & 0.4283 & 0.4300 & 0.4371 & 0.2814 & 0.4593 & 0.2117 & 0.3547 & 0.9021 & 0.8092 & 84.0 \
%     \bottomrule
%     \end{tabular}
%     }
%     \caption{Results of ablation experiments on more models.}
%     \label{table:ablation_study}
% \end{table*}\textbf{}

\section{Needle In A Haystack Test}
\label{appendix:needle_in_a_haysatck}

To ensure that the concepts in our dataset are novel to LLMs, we designed and implemented the "needle in a haystack" test. The purpose of this test is to evaluate the ability of LLMs to retrieve specific information from large corpora, thereby verifying whether the concepts we selected are indeed new to the LLMs.

\subsection{Experimental Setup}

A large corpus was prepared, incorporating text data from diverse online sources such as news articles, blog posts, and social media content. This corpus is intended to simulate the training data typically encountered by LLMs. We selected a set of concepts that were added to our Urban Dictionary dataset after the LLMs' knowledge cutoff date. For each concept, we generated a unique phrase embedding the concept and randomly inserted this phrase into the corpus. This setup was designed to mimic the challenge of locating specific information in a vast dataset. We tested the LLMs' ability to identify and extract each phrase from the corpus using specific prompts. If an LLM successfully retrieved the inserted phrase, the corresponding concept was considered known to the model. If the model failed to find the phrase, the concept was deemed novel.

\begin{figure*}
\centering
\includegraphics[width=0.85\textwidth]{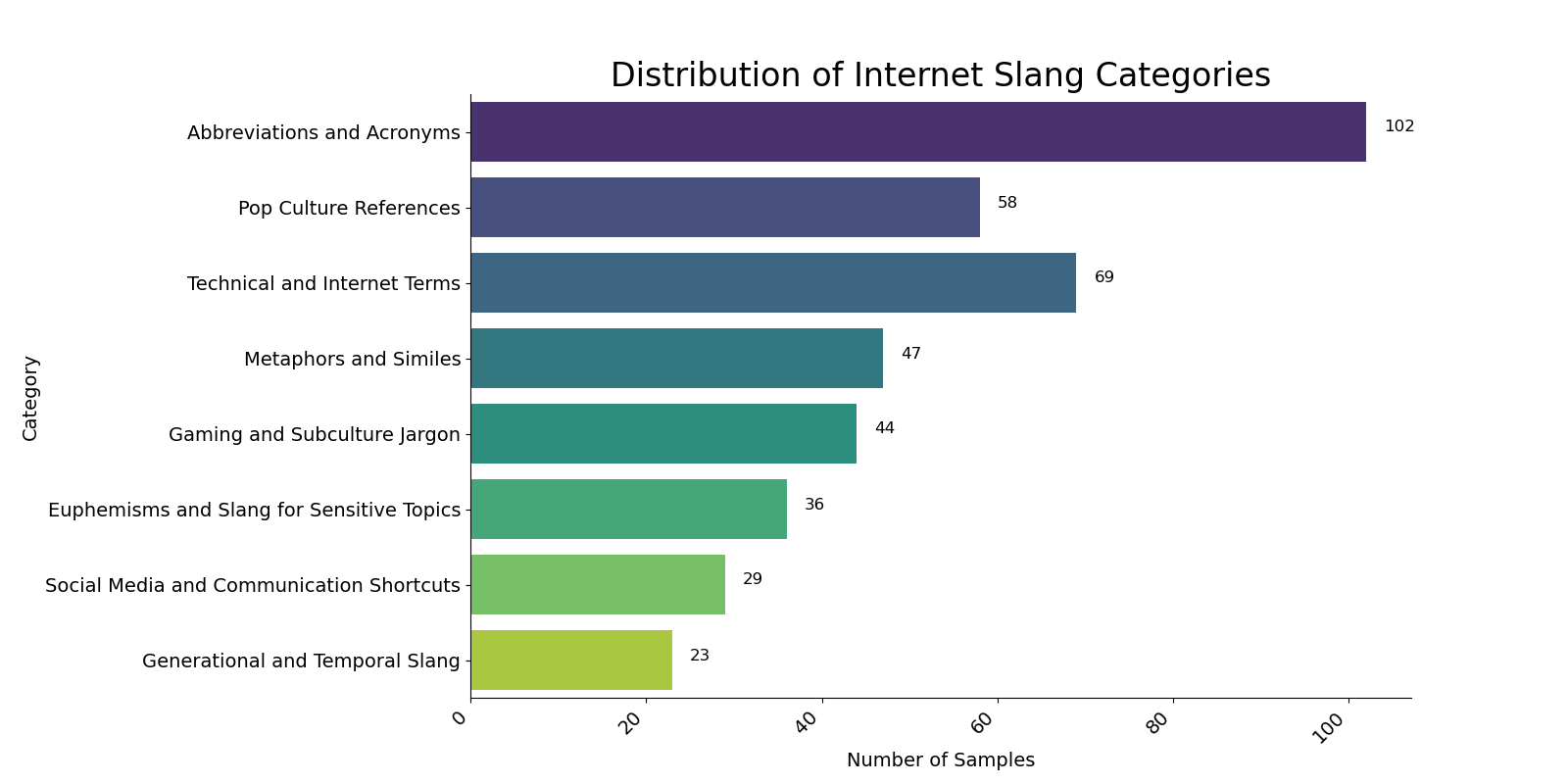}
\caption{Distribution of internet slang categories across our dataset.}
\label{fig:internet_slang_distribution}
\end{figure*}

% Table to show consistency results, with volunteers as the horizontal axis
\begin{table*}[ht]
\centering
\begin{tabular}{c|ccccc|c}
\toprule
\textbf{Volunteers} & \textbf{1} & \textbf{2} & \textbf{3} & \textbf{4} & \textbf{5} & \textbf{Average} \\
\hline
Consistency (\%) & 96.50 & 97.00 & 96.00 & 97.50 & 98.00 & 97.00 \\
\bottomrule
\end{tabular}
\caption{Consistency of the user votes-based filtering strategy across different volunteers.}
\label{table:user_votes_consistency}
\end{table*}

% Table to show recall and precision results
\begin{table}
\centering
\begin{tabular}{c|cc}
\toprule
\textbf{Volunteers} & \textbf{Recall (\%)} & \textbf{Precision (\%)} \\
\hline
Volunteer 1 & 95.00 & 94.12 \\
Volunteer 2 & 97.00 & 93.27 \\
Volunteer 3 & 95.00 & 95.00 \\
Volunteer 4 & 98.00 & 92.45 \\
Volunteer 5 & 99.00 & 91.67 \\
\hline
\textbf{Average} & \textbf{96.80} & \textbf{93.30} \\
\bottomrule
\end{tabular}
\caption{Recall and precision of the user votes-based filtering strategy as validated by human reviewers.}
\label{table:user_votes_rec_prec}
\end{table}

\subsection{Data Format Example}

Here is an illustrative example of the data format used in the "needle in a haystack" test:

\begin{itemize}
    \item Concept: "Tamagotchi effect"
    \item Inserted Phrase: "Jimmy was so upset when his furby died, he obviously was suffering from the tamagotchi effect."
    \item Corpus Sample: "...The stock market showed surprising resilience today. In other news, Jimmy was so upset when his furby died, he obviously was suffering from the tamagotchi effect. Meanwhile, local sports teams are gearing up for the upcoming championships..."
    \item LLM Prompt (w/o few-shot context): "...Identify the phrase from the text that describes a scenario where a person shows emotional distress due to the cessation of function in an electronic device or machine..."
    \item LLM Output: "Jimmy was so upset when his furby died, he obviously was suffering from the tamagotchi effect."
\end{itemize}

If the LLM accurately extracts the inserted phrase, "tamagotchi effect" would be considered known by the LLM. If not, it is marked as a novel concept. This process was repeated for all selected concepts to determine their novelty to the LLMs.

\subsection{Discussion}

It is also important to note that in the context of this experiment, the use of existing filtering strategies based on the cut-off date for GPT models proved to be more stringent than the "needle in a haystack" test. This resulted in no additional data being filtered out by the test since the dataset had already been screened through more conservative criteria \cite{yin-etal-2023-large} and aligned with an earlier knowledge cut-off date. However, this situation is specific to the dataset used in this study, which is a subset tailored to demonstrate the methodology. For broader applications, especially when utilizing our complete open-source dataset which contains over 180,000 entries, the "needle in a haystack" test becomes essential. This is crucial for effectively assessing the novelty of concepts across a more extensive and diverse corpus.

\section{Validation of User-generated Votes}
\label{appendix:validation_user_votes}

To validate the effectiveness of our user votes-based filtering strategy, we conducted an experiment with the assistance of five English-speaking volunteers from English-speaking countries. This experiment was designed to compare the human judgment against the automated user votes-based method, affirming the reliability of user votes as a metric for assessing data quality.

\subsection{Experiment Design}
The experiment engaged five volunteers to review 200 entries from our dataset. These entries included an equal split of 100 entries that had been filtered out and 100 that had been retained by our pre-existing user votes-based filtering method. Volunteers were instructed to make independent filtering decisions for each entry based on our study's quality criteria.

\subsection{Data Collection and Analysis}
We analyzed the decisions of the volunteers to determine the recall and precision of the user votes method against human judgments. Recall measures the proportion of entries that both humans and the automated method agreed should be filtered, while precision assesses the accuracy of the automated method in filtering entries deemed necessary by human reviewers. Additionally, we calculated consistency rates to quantify the agreement between each volunteer’s decisions and the automated method.

\subsection{Results}
The data, as shown in Table \ref{table:user_votes_rec_prec} and \ref{table:user_votes_consistency}, indicate high recall rates (95\% to 99\%) and precision rates (91.67\% to 94.12\%), along with very high consistency rates (96.50\% to 98.00\%). These metrics collectively demonstrate that the user votes-based method is highly effective at mirroring human judgment in filtering decisions. The results underscore the potential of user votes as a reliable indicator of content quality, validating its use as a principal method for data filtering in our study.

\section{Dataset Categorization}
\label{appendix:categorization_slang}

To address potential biases in our dataset that could arise from overrepresentation of certain internet slang categories, such as metaphors, we classified the slang phrases into eight broad categories. We recruited five English-speaking volunteers from English-speaking countries to assist in the categorization of 408 new concepts used in our experiments.

Each volunteer, drawing on their personal experience and familiarity with internet culture, independently categorized each phrase. The categories were as follows:

\begin{itemize}
    \item Abbreviations and acronyms
    \item Pop culture references
    \item Technical and internet terms
    \item Metaphors and similes
    \item Gaming and subculture jargon
    \item Euphemisms and slang for sensitive topics
    \item Social media and communication shortcuts
    \item Generational and temporal slang
\end{itemize}

This collaborative approach was crucial to ensuring that our dataset was not skewed toward any single category of slang, providing a more balanced foundation for analysis. As shown in Figure \ref{fig:internet_slang_distribution}, our dataset features a diverse array of slang expressions.

\paragraph{Concept Editing and Tuning}
Recent advancements in concept editing and tuning have significantly improved the problem-solving capabilities of models across a wide range of fields~\citep{zhang2023lans, zhang2024geoeval, zhang2024fuse, li2024cmmath, chen2023vlp, zhang2024mm}. These methods, which modify the internal structure of large language models (LLMs), are designed to adjust the output based on newly edited knowledge. In particular, many techniques focus on integrating auxiliary networks or tweaking model parameters to guide responses~\citep{meng2022locating, meng2022mass, mitchell2022memory, yao2023editing, bi2024lpnl}. A promising approach in this area is In-Context Editing (ICE)\citep{bi2024adaptive, bi2024decoding, bi2024factualitydecodingfreelunch, bi2024struedit}, which enables models to adapt by utilizing prompts with modified facts and retrieving relevant editing demonstrations from memory. However, hallucinations and safety issues remain significant challenges in tasks related to LLMs\citep{zhang2023sirens, mei2024not, mei2024hiddenguardfinegrainedsafegeneration}.

\section{Summary of Contributions}
\label{appendix:summary_of_contributions}

While our experiments leverage the UrbanDictionary dataset and focus on a subset of popular LLMs, it is crucial to emphasize that the core contributions of this work extend far beyond these specifics. We would like to highlight three key aspects that underscore the broader impact and applicability of our research:

\paragraph{Dataset construction pipeline} This work goes beyond simply providing a static dataset from a single source. Instead, we have developed a comprehensive, open-source toolbox that empowers researchers and practitioners to continuously collect, clean, and process data from a wide range of sources. This toolbox is designed to be highly adaptable, allowing users to easily integrate and analyze data from various platforms and domains, such as social media, online forums, and digital publications. By offering a flexible and extensible framework, our approach ensures the long-term relevance and applicability of the methodology, enabling researchers to keep pace with the ever-evolving landscape of online language. The open-source nature of the toolbox further encourages collaboration and innovation within the research community, as it allows anyone to leverage and build upon our work to process and analyze data from diverse sources, tailoring it to their specific research questions and requirements.

\paragraph{Benchmarking framework} The \textsc{Slang} benchmark is not merely a one-off evaluation limited to the specific datasets and models used in our experiments. Rather, it presents a comprehensive and generalized framework for assessing the adaptability and comprehension capabilities of a wide range of large language models when faced with the challenges of evolving linguistic phenomena. The benchmark is designed to be model-agnostic and can be seamlessly applied to various datasets and architectures, irrespective of their size, domain, or underlying structure, allowing for standardized evaluations and fair comparisons across different settings. By providing a robust and flexible evaluation framework, \textsc{Slang} sets a new standard for assessing the performance of large language models in the face of linguistic change, facilitating the development of language models that can handle the dynamic nature of human language and paving the way for more adaptable and resilient natural language processing systems.

\paragraph{Enhancing LLMs on the fly} The \textsc{Focus} methodology proposed in this work offers a principled approach for improving the ability of LLMs to grasp, interpret, and adapt to emerging linguistic phenomena on the fly, without the need for retraining or relying on Retrieval-augmented generation (RAG) techniques. Our method is model-agnostic, providing a strong basis for anticipating its applicability and benefits across a broader spectrum of LLMs and architectures, without the computational overhead and data requirements associated with retraining or RAG-based approaches. This positions \textsc{Focus} as an efficient and scalable solution for enhancing LLMs' understanding of new concepts. In industrial applications, \textsc{Focus} has the potential to empower businesses across various domains, enabling real-time product recommendation systems, content moderation, sentiment analysis tools, and customer service chatbots that can swiftly adapt to emerging new concepts and terminology. By enhancing enterprises' responsiveness and adaptability, \textsc{Focus} positions itself as a cost-effective solution that can provide a significant competitive advantage in the fast-paced digital market.

\section{Definitions}

\paragraph{Concept} refers to new ideas or phenomena emerging in language due to human activities, particularly on the internet.

\paragraph{Expression} is a specific phrase or term used to convey these concepts.

\paragraph{Deeper meaning} refers to the underlying significance or implications of a concept beyond its literal expression.

\paragraph{Linguistic shift} denotes the gradual incorporation of new concepts into the language, leading to changes over time.

\end{document}